\documentclass[letterpaper]{article}
\usepackage{aaai}
\usepackage{times}
\usepackage{helvet}
\usepackage{courier}
\newtheorem{myDefinition}{Definition}

\usepackage{xspace}
\usepackage{times}
\usepackage{helvet}
\usepackage{courier}
\usepackage{multirow}
\usepackage{epsfig}
\usepackage{epstopdf}
\usepackage{color}
\usepackage{balance}
\usepackage{url}
\usepackage{subfig}
\usepackage{array}
\usepackage{color}

\newcommand{\sectref}[1]{\S\ref{sec-#1}}

\begin{document}
\nocopyright

\title{Automatic Generation of Alternative Starting Positions for \\ Simple Traditional Board Games}

\author{
Umair Z. Ahmed \\ IIT Kanpur  \\ umair@iitk.ac.in \And
Krishnendu Chatterjee \\ IST Austria \\ krishnendu.chatterjee@ist.ac.at \And
Sumit Gulwani \\ Microsoft Research Redmond \\ sumitg@microsoft.com}

\maketitle
\newif\ifciteTech
\citeTechfalse

\newenvironment{myitemize}{\begin{list}{$\bullet$}
{\setlength{\topsep}{1mm}\setlength{\itemsep}{0.25mm}
\setlength{\parsep}{0.1mm}
\setlength{\itemindent}{0mm}\setlength{\partopsep}{0mm}
\setlength{\labelwidth}{15mm}
\setlength{\leftmargin}{4mm}}}{\end{list}}

\renewcommand{\tabcolsep}{0.35em}

\newenvironment{myenumerate}{
\newcounter{lcount}
\begin{list}{\arabic{lcount}.}
{\usecounter{lcount}\setlength{\topsep}{1mm}\setlength{\itemsep}{0.25mm}
\setlength{\parsep}{0.1mm}
\setlength{\itemindent}{0mm}\setlength{\partopsep}{0mm}
\setlength{\labelwidth}{15mm}
\setlength{\leftmargin}{4mm}}}{\end{list}}

\newcommand\ignore[1]{}
\newcommand\checkk[1]{{{\color{red} #1}}}

\providecommand{\e}[1]{\ensuremath{\times 10^{#1}}}
\newcolumntype{C}{>{\centering\let\newline\\\arraybackslash\hspace{0pt}}m{0.75em}}
\newcolumntype{L}{>{\raggedright\let\newline\\\arraybackslash\hspace{0pt}}m{0.75em}}
\newcolumntype{R}{>{\raggedleft\let\newline\\\arraybackslash\hspace{0pt}}m{0.75em}}

\newcommand\XX{\scriptsize {\color{red}X}}
\newcommand\OO{\scriptsize {\color{blue}O}}

\newcommand\ka{k_1}
\newcommand\kb{k_2}
\newcommand{\straa}{\sigma_1}
\newcommand{\strab}{\sigma_2}
\newcommand{\distr}{\mathcal{D}}

\begin{abstract}
Simple board games, like Tic-Tac-Toe and CONNECT-4, play an important role not 
only in the development of mathematical and logical skills, but also in the 
emotional and social development. 
In this paper, we address the problem of generating targeted starting positions 
for such games. 
This can facilitate new approaches for bringing novice players to mastery, and 
also leads to discovery of interesting game variants. 
We present an approach that generates starting states of varying hardness
levels for player~$1$ in a two-player board game, given rules of the board 
game, the desired number of steps required for player~$1$ to win, 
and the expertise levels of the two players.
Our approach leverages symbolic methods and iterative simulation to 
efficiently search the extremely large state space. 
We present experimental results that include discovery of states of varying hardness levels 
for several simple grid-based board games. The presence of such states
for standard game variants like
$4 \times 4$ Tic-Tac-Toe opens up new games to be played
that have never been played as the default start state is heavily biased. 
\end{abstract}

\section{Introduction}
\label{sec-intro}
\ignore{
Two-player graph games are common object of study in formal
methods. They have traditionally been applied to analysis
of reactive systems (such as reactive synthesis and analysis of open
systems~\cite{PR89,AHK02}). For these applications, 
the traditional question that is studied for graph games (with
various winning conditions) is that of existence of optimal
strategies to ensure winning, {\em irrespective of the hardness of the
optimal strategies}. 

In this paper, we study a novel application of graph games, that of
generating alternative starting positions of a certain hardness 
for simple board games such as Tic-Tac-Toe and CONNECT-4. 
The notion of hardness is inherently connected to the expertise levels
of the players; given different expertise levels, the players play
sub-optimally. Thus, one of the novel aspects of our study of graph
games is that we consider different levels of sub-optimal strategies
and generate starting positions where such strategies succeed, 
and starting positions where such strategies fail. 
This advances practical aspects of the theory around graph games. 
Our novel application takes formal methods to interesting societal 
common-person applications, which is good for the field (i.e., instead 
of indirect impact on society through system design and development, there 
is potential of direct impact on society). 

\subsection*{Significance of Board Games}
}
{\em Board games} involve placing pieces on a
pre-marked surface or board according to a set of rules by taking turns. Some of these 
grid-based two-player games like Tic-Tac-Toe and CONNECT-4 have a relatively simple set of rules, 
yet, they are decently challenging for certain age groups. Such games 
have been immensely popular across centuries. 

Studies show that board games can significantly improve a child's mathematical
ability~\cite{ramani08}. Such early differences in mathematical
ability persist into secondary education~\cite{duncan07}.
Board games also assist with emotional 
and social development of a child. 
They instill a competitive desire to master new skills in order to win.
Winning gives a boost to their self confidence. 
Playing a game within a set of rules helps them to 
adhere to discipline in life. They learn social etiquette;
taking turns, and being patient. Strategy is another huge 
component of board games. 
Children learn cause and effect by observing that decisions they make in the beginning of the game
have consequences later on. 

Board games help elderly people stay mentally sharp and less
likely to develop Alzheimer~\cite{alzheimer}. They also hold a great importance in today's digital
society by strengthening family ties. They
bridge the gap between young and old. They bolster the self-esteem of
children who take great pride and pleasure when an elder spends
playing time with them.

\subsection*{Significance of Generating Fresh Starting States}
Board games are typically played with a default start state 
(e.g., empty board in case of Tic-Tac-Toe and CONNECT-4).
However, there are following drawbacks in starting
from the default starting state, which we use to motivate our goals. 

\smallskip\noindent{\bf Customizing hardness level of a start state.}
The default starting state for a certain game, while being unbiased, {\em might not be conducive for a novice
player to enjoy and master the game}.  Traditional board games in particular are easy to learn but difficult to master because these games have intertwined mechanics and force the player to consider far too many possibilities from the standard starting configurations. 
Players can achieve mastery most effectively if complex mechanics can be simplified and learned in isolation.
Csikszentmihalyi's theory of flow~\cite{cflow} suggests that we can keep the learner in a state of maximal
engagement by continually increasing difficulty to match the
learner's increasing skill. Hence, we need an approach that allows generating start states of a specified hardness level. 
This capability can be used to generate a progression of starting states of increasing hardness. This is similar to
how students are taught educational concepts like addition through a progression of increasingly
hard problems~\cite{chi13}. 

\smallskip\noindent{\bf Leveling the playing field.}
 The starting state for commonly played games is mostly unbiased, and hence {\em does not offer a fair experience 
 for players of different skills.}
  The flexibility to start from some other starting state that is 
  more biased towards the weaker player can allow for leveling the playing field and hence 
  a more enjoyable game.
  Hence, we need an approach that takes as input the expertise levels of players and uses that 
  information to associate a hardness level with a state.

\smallskip\noindent{\bf Generating multiple fresh start states.}
A fixed starting state might have a {\em well-known conclusion}. For example, both
  players can enforce a draw in Tic-Tac-Toe while the first player can
  enforce a win in CONNECT-4~\cite{allis1988}, starting from the default empty starting state. 
  Players can {\em memorize certain moves}
  from a fixed starting state and gain undue advantage. 
  Hence, we need an approach that generates multiple start states (of a specified hardness level). 
  This observation has also inspired the design of Chess960~\cite{chess960} (or Fischer Random Chess), which is a variant of chess that employs the same board and pieces as standard chess; 
  however, the starting position of the pieces on the players' home ranks is randomized. The random setup renders the prospect of obtaining an 
  advantage through memorization of opening lines impracticable, compelling players to rely on their talent and creativity. 

\smallskip\noindent{\bf Customizing length of play.}
 People sometimes might be disinterested in playing a game if it
  takes {\em too much time to finish}. However, selecting non-default starting 
  positions allow the potential of a shorter game play. Certain interesting situations might manifest only in states 
that are typically not easily reachable from the start state, or require {\em too many steps}. 
The flexibility to start from such states might lead to more opportunities for 
practice of  specific targeted strategies. Thus, we need an approach that can take as 
input a parameter for the number of steps that can lead to a win for a given player.  

\smallskip\noindent{\bf Experimenting with game variants.}
While people might be hesitant to learn a game with completely different new rules, 
it is quite convenient to change the rules slightly. For example, instead of allowing for
straight-line matches in each of row, column, or diagonal (RCD) in
Tic-Tac-Toe or CONNECT-4, one may restrict the matches to say only row
or diagonal (RD). However, {\em the default starting state of a new game may be heavily biased towards a player}; as a result 
that specific game might not have been popular.  For example, consider the game of Tic-Tac-Toe (3,4,4), where the goal is to make a
straight line of $3$ pieces, but on a $4 \times 4$ board. 
In this game, the person who plays first invariably almost
always wins even with a naive strategy. Hence, such a game has never been popular. 
However, there can be non-default unbiased states for such games and starting from those states 
can make playing such games interesting. Hence, we need an approach that is parameterized by the rules of a game. This also has 
the advantage of experimenting with new games or variants of existing games.
\ignore{In summary, we need an approach to generate {\em multiple} start states of {\em specified hardness levels}, 
given {\em expertise levels of the players} and {\em length of plays}, for traditional {\em board games and their variants}.}

\subsection*{Problem Definition and Search Strategy}
We address the problem of automatically generating 
{\em interesting} starting states (i.e., states of desired hardness levels) for a given two-player board game. 
Our approach takes as input the {\em rules of a board game} (for game variants) and {\em the desired number of steps
required for player 1 to win} (for controlling the length of play). 
It then generates multiple starting states of varying hardness levels 
(in particular, {\em easy, medium, or hard}) for player $1$ for various {\em expertise 
level combinations of the two players}. 
We formalize the exploration of a game as a strategy tree and the expertise level of a player as depth of the strategy tree. The hardness of a state is defined 
w.r.t.~the fraction of times player $1$ will win, while playing a strategy of depth $\ka$ against an opponent who plays a strategy of depth $\kb$. 

Our solution employs a novel combination of symbolic methods 
and iterative simulation to efficiently search for desired states. Symbolic
methods are used to compute the winning set for player $1$. These methods 
work particularly well for navigating a state space where the
transition relation forms a sparse directed acyclic graph (DAG). Such is the case for those board
games in which a piece once placed on the board doesn't move, as in
Tic-Tac-Toe and CONNECT-4. Minimax simulation is used to identify the hardness of a given winning state. 
Instead of randomly sampling the winning set to identify a state of a 
certain hardness level, we identify states of varying hardness levels 
in order of increasing values of $\ka$ and $\kb$. 
The key observation is that hard states are much fewer than 
easy states, and for a given $\kb$, interesting states for higher values 
of $\ka$ are a subset of hard states for smaller values of $\ka$.

\ignore{
\subsection*{Results}
Though our general search methodology applies for any graph game, we focus on 
generating interesting starting states in traditional simple board games and their variants--these are games
whose transition relation forms a sparse DAG (as opposed to an arbitrary graph).
Generating starting states in simple and traditional games, as compared to 
games with complicated rules, is both more challenging and more relevant.
First, in sophisticated games with complicated rules interesting states are 
likely to be abundant and hence easier to find, whereas finding interesting 
states in simple games is more challenging.
Second, games with complicated rules are hard to learn, whereas simple 
variants of traditional games (such as larger or smaller board size, or 
changing winning conditions from RCD to RD) are easier to adopt.
Hence finding interesting start states in traditional games and its variants 
is the more relevant and challenging question that we address in this work. 
We thus present a framework to easily describe
new board games like Tic-Tac-Toe or CONNECT-4 or their variants. Our implementation works for 
games that can be described using this framework. 

We experimented with Tic-Tac-Toe, CONNECT, Bottom-2 (a new game that is a 
hybrid of Tic-Tac-Toe and CONNECT) games and several of their variants like RD or RC as winning rules instead of RCD.
We were able to generate several (tens of) starting states of various hardness levels for various expertise 
levels and number of winning steps. 
Two important findings of the experiments are: (i)~discovery of starting states of various hardness levels 
in  these traditional board games, and furthermore discovering them in games such as Tic-Tac-Toe on 
$4\times 4$ board size where the default start state is heavily biased; and 
(ii)~these states are rare and thus require a non-trivial search strategy like ours to find them.

}

\subsection*{Contributions}
\begin{myitemize}

\item We introduce and study a novel aspect of graph games, namely 
  generation of starting states. In particular, we address the problem of 
  generating starting states of varying hardness levels parameterized by 
  look-ahead depth of the strategies of the two players, 
  the graph game description, and the number of steps
  required for winning (\sectref{formalism}).

\item We present a novel search methodology for generating desired initial states. It involves 
  combination of symbolic methods and iterative simulation to
  efficiently search a huge state space (\sectref{methodology}). 
\item We present experimental results that illustrate the
  effectiveness of our search methodology (\sectref{results}). We 
  produce a collection of initial states of varying hardness
  levels for standard games as well as their variants (thereby discovering some interesting variants of the
  standard games in the first place). 
\end{myitemize}

\ignore{Though our general search methodology applies for any graph game, 
for experimental results we focus on generating interesting starting states 
in traditional simple board games and their variants, as compared to 
games with complicated rules, as such games 
are hard to learn. 
In contrast, simple variants of traditional games (such as larger 
or smaller board size, or changing winning conditions from RCD to RD) are easier 
to adopt.
Hence finding interesting start states in traditional games and its variants 
is the more relevant question that we address in this work. }

While our search methodology applies to any graph game; in 
our experiments we focus on generating starting states in {\em simple} 
board games and their variants as opposed to games with complicated rules.
Games with complicated rules are hard to learn, whereas simple  
variants of traditional games are easier to adopt. 
The problem of automated generation of starting states should also be experimented for complex games as future work; however, given that 
this problem has not even been studied for simple games in the past, and 
our approach makes valuable discovery for simple games, is a key contribution 
of this paper.

\ignore{This paper is organized as follows. We start out with a formal background of graph 
games and then formally state our problem definition in \sectref{formalism}. 
We present our search methodology for generating desired initial states in \sectref{methodology}. 
We then present a framework for describing (rules of) new board games that 
are like Tic-Tac-Toe or CONNECT-4 (or simple variants of those) in \sectref{framework}.
We describe experimental results for several instantiations of this framework in \sectref{results}. 
We describe related work in \sectref{related} and then conclude in \sectref{concl}.}

\newcommand{\epre}{\mathsf{EPre}}
\newcommand{\apre}{\mathsf{APre}}
\newcommand{\post}{\mathsf{Post}}
\newcommand{\set}[1]{\{ #1 \}}
\newcommand{\seq}[1]{\langle #1 \rangle}
\newcommand{\CuDD}{{\sc CuDD}}
\newcommand\textt[1]{\mbox{#1}}

\section{Problem Definition}
\label{sec-formalism}

\subsection{Background on Graph Games}
\label{sec-background}
\noindent{\bf Graph games.} An alternating graph game (for short,
graph game) $G=((V,E),(V_1,V_2))$ consists of a finite graph $G$ with vertex 
set $V$, a partition of the vertex set into player-1 vertices $V_1$ and player-2 
vertices $V_2$, and edge set $E \subseteq ((V_1 \times V_2) \cup (V_2 \times V_1))$. 
The game is alternating in the sense that the edges of player-1 vertices go to 
player-2 vertices and vice-versa.
The game is played as follows: the game starts at a starting vertex $v_0$;
if the current vertex is a player-1 vertex, then player~1 chooses an 
outgoing edge to move to a new vertex; if the current vertex is a player-2
vertex, then player~2 does likewise.
The winning condition is given by a target set $T_1 \subseteq V$ 
for player~1; and similarly a target set $T_2 \subseteq V$ for player~2.
If the target set $T_1$ is reached, then player~1 wins; 
if $T_2$ is reached, then player~2 wins;
else we have a draw.

\smallskip\noindent{\bf Examples.} 
The class of graph games provides the mathematical framework to study many 
board games like Chess or Tic-Tac-Toe.
For example, in Tic-Tac-Toe the vertices of the graph represent 
the board configurations and whether it is player~1 ($\times$) or player~2 
($\circ$) to play next. The set $T_1$ (resp. $T_2$) is the set of board 
configurations with three consecutive $\times$ (resp. $\circ$) 
in a row, column, or diagonal.

\smallskip
\noindent{\bf Classical game theory result.}
A classic result in the theory of graph games~\cite{GS53} shows that for every graph game
with respective target sets for both players,
from every starting vertex one of the following three conditions hold:
(1)~player~1 can enforce a win no matter how player~2 plays (i.e., there 
is a strategy for player~1 to play to ensure winning against all possible 
strategies of the opponent);
(2)~player~2 can enforce a win no matter how player~1 plays;
or
(3)~both players can enforce a draw 
(player~1 can enforce a draw no matter how player~2 plays, and 
player~2 can enforce a draw no matter how player~1 plays).
The classic result (aka determinacy) rules out the following 
possibility: 
against every player-1 strategy, player~2 can win;
and 
against every player-2 strategy,  player~1 can win.
In the mathematical study of game theory, the theoretical question (which 
ignores the notion of hardness) is as follows: 
given a designated starting vertex $v_0$ determine whether case~(1),
case~(2), or case~(3) holds.
In other words, the mathematical game theoretic question concerns 
the best possible way for a player to play to ensure the best possible result.
The set $W_j$ is defined as the set of vertices such that player~1 can ensure
to win within $j$-moves; and the winning set $W^1$ of vertices of player~1 is 
the set $\bigcup_{j\geq 0} W_j$ where player~1 can win in any number of moves.
Analogously, we define $W^2$; and the classical game theory question is 
formally stated as follows:
given a designated starting vertex $v_0$ decide whether $v_0$ belongs to $W^1$
(player-1 winning set) or to $W^2$ (player-2 winning set) or 
to $V \setminus (W^1 \cup W^2)$ (both players draw ensuring set).

\subsection{Formalization of Problem Definition}
\label{sec-probdef}

\smallskip\noindent{\bf Notion of hardness.}
The mathematical game theoretic question ignores two aspects. 
(1)~The notion of hardness:  It is concerned with optimal
strategies irrespective of hardness;
and (2)~the problem of generating different starting vertices.
We are interested in generating starting vertices of different
hardness. 
The hardness notion we consider is the depth of the tree a player can 
explore, which is standard in artificial intelligence. 

\smallskip
\noindent{\bf Tree exploration in graph games.} 
Consider a player-1 vertex $u_0$. 
The \emph{search tree} of depth~1 is as follows: 
we consider a tree rooted at $u_0$ such that children of $u_0$ are the vertices 
$u_1$ of player~2 such that $(u_0,u_1)\in E$ (there is an edge from $u_0$ to 
$u_1$); and for every vertex $u_1$ (that is a children of $u_0$) the children
of $u_1$ are the vertices $u_2$ such that $(u_1,u_2) \in E$, and they are
the leaves of the tree.
This  gives us the search tree of depth~1, which intuitively corresponds to 
exploring one round of the play.
The search tree of depth $k+1$ is defined inductively from the search tree of 
depth $k$, where we first consider the search tree of depth~1 and replace
every leaf by a search tree of depth $k$. 
The depth of the search tree denotes the depth of reasoning (analysis depth) 
of a player. 
The search tree for player~2 is defined analogously.

\smallskip
\noindent{\bf Strategy from tree exploration.} 
A \emph{depth-$k$ strategy} of a player that does a tree exploration of 
depth $k$ is obtained by the \emph{classical min-max} reasoning 
(or backward induction) on the search tree.
First, for every vertex $v$ of the game we associate a number (or reward) 
$r(v)$ that denotes how favorable is the vertex for a player to win.
Given the current vertex $u$, a depth-$k$ strategy is defined as 
follows:
first construct the search tree of depth $k$ and evaluate the tree
bottom-up with min-max reasoning. 
In other words, a leaf vertex $v$ is assigned reward $r(v)$, where the 
reward function $r$ is game specific, and intuitively,  $r(v)$ denotes how 
``close" the vertex $v$ is to a winning vertex (see the following paragraph
for an example). 
For a vertex in the tree if it is a player-1 (resp.
player-2) vertex we consider its reward as the maximum (resp. minimum)
of its children, and finally, for vertex $u$ (the root) the strategy 
chooses \emph{uniformly at random} among its children with the highest reward.
Note that the rewards are assigned to vertices only based on the vertex
itself without any look-ahead, and the exploration 
is captured by the classical min-max tree exploration.

\smallskip
\noindent{\bf Example description of tree exploration.}
Consider the example of the Tic-Tac-Toe game. 
We first describe how to assign reward $r$ to board positions.
Recall that in the game of Tic-Tac-Toe the goal is to form a line of three 
consecutive positions in a row, column, or diagonal.
Given a board position, (i)~if it is winning for player~1, then 
it is assigned reward $+\infty$; (ii)~else if it is winning for player~2,
then it is assigned reward $-\infty$; (iii)~otherwise it is assigned 
the score as follows: let $n_1$ (resp. $n_2$) be the number of two consecutive 
positions of marks for player~1 (resp. player~2) that can be extended to satisfy
the winning condition.
Then the reward is the difference $n_1-n_2$.
Intuitively, the number $n_1$ represents the number of possibilities for 
player~1 to win, and $n_2$ represents the number of possibilities for player~2,
and their difference represents how favorable the board position is for player~1. 
If we consider the depth-1 strategy, then the strategy chooses all board
positions uniformly at random; a depth-2 strategy chooses the center and 
considers all other positions to be equal;
a depth-3 strategy chooses the center and also recognizes that the next best 
choice is one of the four corners.
An illustration is given in the
\ifciteTech{appendix (attached as supplementary material).
}
\else{appendix.}
\fi
This example illustrates that as the depth increases, the strategies become
more intelligent for the game.

\smallskip
\noindent{\bf Outcomes and probabilities given strategies.}
Given a starting vertex $v$, a depth-$\ka$ strategy $\sigma_1$ for player~1, and 
depth-$\kb$ strategy $\sigma_2$ for player~2, let $O$ be the set of possible 
outcomes, i.e., the set of possible plays given $\sigma_1$ and $\sigma_2$ 
from $v$, where a play is a sequence of vertices.  
The strategies and the starting vertex define a probability distribution over
the set of outcomes which we denote as $\Pr_v^{\sigma_1,\sigma_2}$, i.e., 
for a play $\rho$ in the set of outcomes $O$ we have 
$\Pr_v^{\sigma_1,\sigma_2}(\rho)$ is the probability of $\rho$ given the 
strategies.
Note that strategies are randomized (because strategies choose distributions
over the children 
in the search tree exploration), and 
hence define a probability distribution over the set of outcomes.
This probability distribution is used to formally define the
notion of hardness we consider.

\smallskip
\noindent{\bf Problem definition.} 
We consider several board games (such as Tic-Tac-Toe,
CONNECT-4, and variants), and 
our goal is to obtain starting positions that are of different hardness levels,
where our hardness is characterized by strategies of different depths. Precisely, consider a depth-$\ka$ strategy for player~1, and 
depth-$\kb$ strategy for player~2, and a starting vertex $v \in W_j$ that is
winning for player~1 within $j$-moves and a winning move 
(i.e., $j+1$ moves for player~1 and $j$ moves of player~2).
We classify the starting vertex as follows:
if player~1 wins (i)~at least $\frac{2}{3}$ times, then we call it~\emph{easy (E)};
(ii)~at most $\frac{1}{3}$ times, then we call it~\emph{hard (H)}; 
(iii)~otherwise \emph{medium (M)}.

\begin{myDefinition}{\bf ($(j,\ka,\kb)$-Hardness).}
Consider a vertex $v \in W_j$ that is winning for player~1 within $j$-moves.
Let $\sigma_1$ and $\sigma_2$ be a depth-$\ka$ strategy for player~1 and 
depth-$\kb$ strategy for player~2, respectively.
Let $O_1 \subseteq O$ be the set of plays that belong to the set of outcomes and 
is winning for player~1.
Let 
$\Pr_v^{\sigma_1,\sigma_2}(O_1)=\sum_{\rho\in O_1}\Pr_v^{\sigma_1,\sigma_2}(\rho)$ 
be the probability of the winning plays.
The $(\ka,\kb)$-classification of $v$ is:
(i)~if $\Pr_v^{\sigma_1,\sigma_2}(O_1) \geq \frac{2}{3}$, then $v$ is easy (E);
(ii)~if $\Pr_v^{\sigma_1,\sigma_2}(O_1) \leq \frac{1}{3}$, then $v$ is hard (H);
(iii)~otherwise it is medium (M).
\end{myDefinition}

\noindent{\em Remark~1.}
In the definition above we chose the probabilities $\frac{1}{3}$ and 
$\frac{2}{3}$, however, the probabilities in the definition could be easily 
changed and experimented.
We chose $\frac{1}{3}$ and $\frac{2}{3}$ to divide the interval $[0,1]$ 
symmetrically in regions of E, M, and H. 
In this work, we present result based on the above definition.

Our goal is to consider various games and identify vertices
of different categories (hard for depth-$\ka$ vs.~depth-$\kb$, 
but easy for depth-($\ka$$+$$1$) vs.~depth-$\kb$, for small $\ka$ and $\kb$).

\noindent{\em Remark~2.}
In this work we consider classical min-max reasoning for tree exploration.
A related notion is Monte Carlo Tree Search (MCTS) which in general converges 
to min-max exploration, but can take a long time. 
However, this convergence is much faster in our setting, since we consider 
simple games that have great symmetry, and explore only small-depth strategies.

\section{Search Strategy}
\label{sec-methodology}

\subsection{Overall methodology}
\noindent{\bf Generation of $j$-steps win set.}
Given a game graph $G=((V,E),(V_1,V_2))$ along with target sets $T_1$ and 
$T_2$ for player~1 and player~2, respectively, our first goal is to compute the 
set of vertices  $W_j$ such that player~1 can win within $j$-moves.
For this we define two kinds of predecessor operators: one predecessor operator 
for player~1, which uses existential quantification over successors, and one for 
player~2, which uses universal quantification over successors.  
Given a set of vertices $X$, let 
$\epre(X)$ (called existential predecessor) denote the set of player-1 vertices  
that has an edge to $X$; i.e., $\epre(X)= \set{u \in V_1 \mid \textt{ there exists } 
v \in X \textt{ such that } (u,v) \in E}$ (i.e., player~1
can ensure to reach $X$ from $\epre(X)$ in one step);
and 
$\apre(X)$ (called universal predecessor) denote the set of player-2 vertices that has 
all its outgoing edges to $X$; i.e., $\apre(X)= \set{u \in V_2 \mid \textt{ for all }
(u,v) \in E \textt{ we have } v \in X}$ (i.e., irrespective of the choice
of player~2 the set $X$ is reached from $\apre(X)$ in one step).
The computation of the set $W_j$ is defined inductively:
$W_0=\epre(T_1)$ (i.e., player~1 wins with the next move to reach $T_1$);
and $W_{i+1} = \epre(\apre(W_i))$.
In other words, from $W_i$ player~1 
can win within $i$-moves, and from $\apre(W_i)$ irrespective
of the choice of player~2 the next vertex is in $W_i$; and hence 
$\epre(\apre(W_i))$ is the set of vertices such that~player~1 can win 
within $(i+1)$-moves.

\smallskip
\noindent{\bf Exploring vertices from $W_j$.} 
The second step is to explore vertices from $W_j$, for increasing values
of $j$ starting with small values of $j$. 
Formally, we consider a vertex $v$ from $W_j$, consider a depth-$\ka$
strategy for player~1 and a depth-$\kb$ strategy for player~2, and 
play the game multiple times with starting vertex $v$ to find out 
the hardness level with respect to $(\ka,\kb)$-strategies, i.e., the 
$(\ka,\kb)$-classification of $v$. 
Note that from $W_j$ player~1 can win within $j$-moves. Thus
the approach has the benefit that player~1 has a winning strategy
with a small number of moves and the game need not be played
for long.

\smallskip
\noindent{\bf Two key issues.} There are two main computational 
issues associated with the above approach in practice. 
The first issue is related to the size of the state space (number of 
vertices) of the game which makes enumerative approach to analyze the game 
graph explicitly computationally infeasible. 
For example, the size of the state space of 
Tic-Tac-Toe $4\times4$ game is 6,036,001; and a 
CONNECT-4  $5\times 5$ game is 69,763,700 (above 69 million).
Thus any enumerative method would not work for such large game graphs.
The second issue is related to exploring the vertices from $W_j$. If $W_j$ 
has a lot of witness vertices, then playing the game multiple times from 
all of them will be computationally expensive. 
So we need an initial metric to guide the search of vertices
from $W_j$ such that the metric computation is 
inexpensive.
We solve the first issue with \emph{symbolic} methods, and the second 
one by \emph{iterative simulation}.

\subsection{Symbolic methods}
We discuss the symbolic methods to analyze games with large state spaces. 
The key idea is to represent the games symbolically (not 
with explicit state space) using variables, and 
operate on the symbolic representation.
The key object used in symbolic representation are called 
BDDs (boolean decision diagrams)~\cite{Bryant86} that can efficiently 
represent a set of vertices using a DAG representation of a boolean
formula representing the set of vertices.
The tool \CuDD\ supports many symbolic representation of state
space using BDDs and supports many operations on symbolic 
representation on graphs using BDDs~\cite{CuDD}.

\smallskip
\noindent{\bf Symbolic representation of vertices.}
In symbolic methods, a game graph is represented by a 
set of variables $x_1,x_2,\ldots,x_n$ such that each of
them takes values from a finite set (e.g., $\times,\circ$, 
and blank symbol); and each vertex of the game
represents a valuation assigned to the variables.
For example, the symbolic representation of the game of
Tic-Tac-Toe of board size $3\times 3$ consists of 
ten variables $x_{1,1},x_{1,2}, x_{1,3},x_{2,1}\ldots,x_{3,3},x_{10}$, 
where the first nine variables $x_{i,\ell}$ denote the symbols in the 
board position $(i,\ell)$ and the symbol is either $\times, \ \circ,$ or 
blank; 
and the last variable $x_{10}$ denotes whether it is player~1 or 
player~2's turn to play.
Note that the vertices of the game graph not only contains the information 
about the board configuration, but also additional information such as the 
turn of the players.
To illustrate how a symbolic representation is efficient, consider the 
set of all valuations to boolean variables $y_1, y_2, \ldots, y_n$ 
where the first variable is true, and the second variable is false: 
an explicit enumeration requires to list $2^{n-2}$ valuations, where as a 
boolean formula representation is very succinct.
Symbolic representation with BDDs exploit such succinct representation
for sets of vertices, and are used in many applications, e.g. hardware 
verification~\cite{Bryant86}.

\smallskip
\noindent{\bf Symbolic encoding of transition function.}
The transition function (or the edges) are also encoded in a 
symbolic fashion: instead of specifying every edge, the symbolic
encoding allows to write a program over the variables 
to specify the transitions.
The tool \CuDD\ takes such a symbolic description written as a 
program over the variables and constructs a BDD representation 
of the transition function.
For example, for Tic-Tac-Toe, a program to
describe the symbolic transition is: 
the program maintains a set $U$ of positions of the board that
are already marked; and at every point receives an input $(i,\ell)$ 
from the set $\set{(a,b) \mid 1 \leq a,b \leq 3} \setminus U$ of 
remaining board positions from the player of the current turn; 
then adds $(i,\ell)$ to the set $U$ and sets the variable $x_{i,\ell}$ as 
$\times$ or $\circ$ (depending on whether it was player~1 or
player~2's turn).
This gives the symbolic description of the transition function.


\smallskip
\noindent{\bf Symbolic encoding of target vertices.}
The set of target vertices is encoded as a boolean formula
that represents a set of vertices.
For example, in Tic-Tac-Toe the set of target vertices
for player~1 is given by the following boolean formula:
\begin{footnotesize}
\[
\begin{array}{l}
\exists i,\ell. \ 1 \leq i, \ell \leq 3. \  
(x_{i,\ell}= \times \land x_{i+1,\ell}=\times \land x_{i+2,\ell}= \times )
\\[1ex]
\lor (x_{i,\ell}= \times \land x_{i,\ell+1}=\times \land x_{i,\ell+2}=\times ) \\[1ex]
\lor (x_{2,2}= \times \land \\ \indent \!((x_{1,1}= \times \land x_{3,3}=\times) \lor \!(x_{3,1}= \times \land x_{1,3}=\times)\! )\!
 ) \\[1ex]
\land  \ \textt{Negation of above with $\circ$ to specify player~2 not winning}
\end{array}
\]
\end{footnotesize}
The above formula states that either there is some column 
($x_{i,\ell},  x_{i+1,\ell}$ and $x_{i+2,\ell}$) that is winning for 
player~1; 
or a row ($x_{i,\ell}, x_{i,\ell+1}$ and $x_{i,\ell+2}$) that is winning for 
player~1; or there is a diagonal ($x_{1,1}, x_{2,2}$ and $x_{3,3}$; 
or $x_{3,1}, x_{2,2}$ and $x_{1,3}$)
that is winning for player~1; and player~2 has not won already.
To be precise, we also need to consider the BDD that represents all
valid board configurations (reachable vertices from the empty board) 
and intersect the BDD of the above formula with valid board
configurations to obtain the target set $T_1$.

\smallskip
\noindent{\bf Symbolic computation of $W_j$.}
The symbolic computation of $W_j$ is as follows: given 
the boolean formula for the target set $T_1$ we obtain the
BDD for $T_1$; and the \CuDD\ tool supports both $\epre$
and $\apre$ 
as basic operations using symbolic functions;
i.e., the tool takes as input a BDD representing a set $X$
and supports the operation to return the BDD for $\epre(X)$
and $\apre(X)$.
Thus we obtain the symbolic computation of $W_j$.

\subsection{Iterative simulation}
We now describe a computationally inexpensive 
way to aid sampling of vertices as candidates for 
starting positions of a given hardness level.
Given a starting vertex $v$, a depth-$\ka$ strategy for player~1, 
and a depth-$\kb$ strategy for player~2, we need to consider the tree 
exploration of depth $\max\set{\ka,\kb}$ to obtain the hardness of $v$.
Hence if either of the strategy is of high depth, then it is
computationally expensive. 
Thus we need a preliminary metric that can be computed relatively 
easily for small values of $\ka$ and $\kb$ as a guide for vertices to be 
explored in depth.
We use a very \emph{simple metric} for this purpose. 
The hard vertices are rarer than the easy vertices,
and thus we rule out easy ones quickly using the following approach: \\
\underline{{\em If $\ka$ is large}}: 
Given a strategy of depth $\kb$, the set of hard vertices for higher values 
of $\ka$ are a subset of the hard vertices for smaller values of $\ka$.
Thus we iteratively start with smaller values and proceed 
to higher values of $\ka$ only for vertices that are already hard for 
smaller values of $\ka$. \\
\underline{{\em If $\kb$ is large}}:  
Here we exploit the following intuition. 
Given a strategy of depth $\ka$, a vertex which is hard for high value of 
$\kb$ is likely to show indication of hardness already in small values
of $\kb$. 
Hence we consider the following approach.
For the vertices in $W_j$, we fix a depth-$\ka$ strategy, and fix a small 
depth strategy for the opponent and assign the vertex a number (called {\em score})
based on the performance of the depth-$\ka$ strategy and the small depth 
strategy of the opponent.
The score indicates the fraction of games won by the depth-$\ka$ strategy
against the opponent strategy of small depth. 
The vertices that have low score 
are then 
iteratively simulated against depth-$\kb$ strategies of the opponent 
to obtain vertices of different hardness level.
This heuristic serves as a simple metric to explore vertices
for large value of $\kb$ starting with small values of $\kb$.

\section{Framework for Board Games}
\label{sec-framework}
We now consider the specific problem of board games.
We describe a framework to specify several variants of
two-player grid-based board games such as Tic-Tac-Toe, CONNECT-4.

\smallskip
\noindent{\bf Different parameters.}
Our framework allows three different parameters
to generate variants of board games.
(1)~The first parameter is the \emph{board size}; e.g., the board size could 
be $3\times 3$; or $4\times 4$; or $4 \times 5$ and so on.
(2)~The second parameter is the way to specify the winning condition,
where a player wins if a sequence of the player's moves are in a line, which could be along a row (R), a column (C), or the diagonal (D). 
The user can specify any combination: 
(i)~RCD (denoting the player 
wins if the moves are in a line along a row, column or diagonal);
(ii)~RC (line must be along a row or column, but diagonal lines are 
not winning);
(iii)~RD (row or diagonal, not column); or
(iv)~CD (column or diagonal, not row).
(3)~The third parameter is related to the allowed moves of the 
player. At any point the players can choose any available column (i.e., column with at least one empty position) but can be restricted according to the following 
parameters: (i)~Full gravity (once a player chooses a column, the move
is fixed to be the lowest available position in that column);
(ii)~partial gravity-$\ell$ (once a player chooses a column, the move can be one of the bottom-$\ell$ available positions in the 
column);
or (iii)~no gravity (the player can choose any of the available positions
in the column).
Observe that Tic-Tac-Toe is given as (i)~board size 
$3\times 3$; (ii)~winning condition RCD; and (iii)~no-gravity;
whereas in CONNECT-4 the winning condition is still RCD but moves
are with full gravity.
But in our framework there are many new variants of the previous 
classical games, e.g., Tic-Tac-Toe in a board of size
$4\times 4$ but diagonal lines are not winning (RC);
and Bottom-2 (partial gravity-2) which is between Tic-Tac-Toe and CONNECT
games in terms of moves allowed.

\smallskip
\noindent{\bf Features of our implementation.} 
We have implemented our approach 
and the main features that our implementation supports are:
(1)~Generation of starting vertices of different hardness level
if they exist. 
(2)~Playing against opponents of different levels. 
We have implemented the depth-$\kb$ strategy of the opponent 
for $\kb=1,2$ and $3$ (typically in all the above games 
depth-$3$ strategies are quite intelligent, and hence we do not explore 
larger values of $\kb$).
Thus, a learner (beginner) can consider starting with 
board positions of various hardness levels and play with opponents 
of different skill level and thus hone her ability to play the 
game and be exposed to new combinatorial challenges of the game.

\begin{table}
\centering
{\scriptsize
\caption{\small CONNECT-3 \& -4 against depth-$3$ strategy of opponent;
(C-3 (resp. C-4) stands for CONNECT-3 (resp. CONNECT-4)).
The third column ($j$) denotes whether we explore from $W_2$ or $W_3$.
The sixth column denotes sampling to select starting vertices 
if $|W_j|$ is large: ``All'' denotes that we explore all vertices in $W_j$, 
and Rand denotes first sampling 5000 vertices randomly from $W_j$ and 
exploring them.
The E, M, and H columns give the number of easy, medium, or hard vertices among 
the sampled vertices.
For each $\ka=1,2,$ and $3$ the sum of E, M, and H columns is equal
to the number of sampled vertices, and * denotes the number of remaining vertices.
Observe that $|W_j|$ is small fraction of $|V|$ (this illustrates the significance of our use of symbolic methods
as opposed to the prohibitive explicit enumerative search). Also, observe that vertices labeled medium and hard are a small 
fraction of the sampled vetices (this illustrates the significance of our efficient iterative sampling strategy).
} 
\label{tabConnect} 
\scalebox{0.8}{
\begin{tabular}{|*{6}c|*{3}c|*{3}c|*{3}c|} \hline
Game & State & $j$ & {Win} & {No. of} & Sampling & \multicolumn{9}{c|}{$\kb=3$}   \\ \cline{7-15}
{} & {Space} & {} & {Cond} & {States} & {} & \multicolumn{3}{c|}{$\ka$  = 1} & \multicolumn{3}{c|}{$\ka$  = 2} & \multicolumn{3}{c|}{$\ka$  = 3} \\ 
{} & {$|V|$} & &  & {$|W_j|$} & {} & E&M&H & E&M&H & E&M&H  \\ \hline

{C-3}	& {4.1\e4} & 2 & RCD & 110 & All & *&24&5  & *&3 &0  & *&0&0 \\ 
{4x4}	& {6.5\e4} &   & RC  & 200 & All & *&39&9  & *&23&5  & *&0&0 \\ 
{}		& {7.6\e4} &   & RD  & 418 & All & *&36&17 & *&25&4  & *&0&0 \\ 
{}		& {6.5\e4} &   & CD  & 277 & All & *&41&24 & *&27&21 & *&0&0 \\ \hline 

{C-3} & & 3 & RCD, RC, CD 	& 0  & -   & 	 &&& 	 &&&    && \\ 
{4x4} & &   & RD 			& 18 & All & *&0&0 & *&0&0 & *&0&0 \\ \hline

{C-4} 	& 6.9\e7  & 2 & RCD & 1.2\e6 & Rand & *&184&215 & *&141&129 & *&0&0 \\ 
{5x5} 	& {8.7\e7}&   & RC  & 1.6\e6 & Rand & *&81&239  & *&70&186  & *&0&0 \\ 
{}		& {1.0\e8}&   & RD  & 1.1\e6 & Rand & *&106&285 & *&151&82  & *&0&0 \\ 
{} 		& {9.5\e7}&   & CD  & 5.3\e5 & Rand & *&364&173 & *&209&96  & *&0&0 \\ \hline 

{C-4}&& 3 & RCD & 2.8\e5 & Rand & *&445&832  & *&397&506 & *&208&211 \\ 
{5x5}&&   & RC  & 7.7\e5 & Rand & *&328&969  & *&340&508 & *&111&208 \\
{}	 &&   & RD  & 8.0\e5 & Rand & *&398&1206 & *&464&538 & *&179&111 \\
{} 	 &&   & CD  & 1.5\e5 & Rand & *&146&73   & *&171&110 & *&120&72  \\ \hline
\end{tabular} 
}}
\end{table}

\section{Experimental Results}
\label{sec-results}
Our experiments reveal useful discoveries. 
The main aim 
is to investigate the 
existence of interesting starting vertices and their abundance 
in CONNECT, Tic-Tac-Toe, and Bottom-2 games, for various combinations of
expertise levels and winning rules (RCD, RC, RD, and CD),
for small lengths of plays. 
Moreover, the computation time should be reasonable.
\ignore{
Our key findings show that such vertices exist (but in most cases are rare, 
and thus their discovery is an important finding and illustrates the significance of our non-trivial search strategy) for Tic-Tac-Toe for depth-1 strategies, 
for Bottom-2 for depth-1 and depth-2 strategies, and in CONNECT-4 for depth-1, 
depth-2 and depth-3 strategies, against a depth-3 strategy of the opponent.
\checkk{We also obtain similar results against depth-2 strategy of the opponent.} 
Furthermore, we observe the existence of interesting vertices in 
Tic-Tac-Toe games and its variants over $4\times 4$ board size, where 
the default start vertex is uninteresting.
We next briefly detail our experimental results and important findings 
and finally we show some example board positions.
}




\begin{table}
\centering
{\scriptsize

\caption{\small Bottom-2 against depth-$3$ strategy of opponent.} 
\label{tabBottom2} 
\scalebox{0.96}{
\begin{tabular}{|*{6}c|*{3}c|*{3}c|*{3}c|} \hline
Game & State & $j$ & {Win} & {No. of} & Sampling & \multicolumn{9}{c|}{$\kb=3$}   \\ \cline{7-15}
{} & {Space} & {} & {Cond} & {States} & {} & \multicolumn{3}{c|}{$\ka$  = 1} & \multicolumn{3}{c|}{$\ka$  = 2} & \multicolumn{3}{c|}{$\ka$  = 3} \\ 
{} & {$|V|$} & &  & {$|W_j|$} & {} & E&M&H & E&M&H & E&M&H  \\ \hline

{3x3} & {4.1\e3} & $2$ & RCD & 20 & All & *&5&0 & *&1&0 & *&0&0	\\ 
{} 	  & {4.3\e3} & {}  & RC  & 0  & -   & &&  &    && &   &&	\\ 
{} 	  & {4.3\e3} & {}  & RD  & 9  & All & *&2&1 & *&3&0 & *&0&0	\\ 
{} 	  & {4.3\e3} & {}  & CD  & 1  & All & *&0&0 & *&0&0 & *&0&0	\\ \hline 

{3x3} &  & $3$ & Any & 0 & -  & &&  & && & && \\ \hline

{4x4} & {1.8\e6} & $2$ & RCD & 193  & All & *&12&26   & *&0&2    & *&0&0 \\ 
{} 	  & {2.4\e6} & {}  & RC  & 2709 & All & *&586&297 & *&98&249 & *&0&0 \\ 
{} 	  & {2.3\e6} & {}  & RD  & 2132 & All & *&111&50  & *&18&16  & *&0&0 \\ 
{} 	  & {2.4\e6} & {}  & CD  & 1469 & All & *&123&53  & *&25&8   & *&0&0 \\ \hline 

{4x4} && $3$ & RCD & 0  & -   &    &&   &  &&  &   &&  \\ 
{} 	  && {}  & RC  & 90 & All & *&37&31 & *&0&0 & *&0&0 \\
{}    && {}  & RD  & 24 & All & *&1&2   & *&0&0 & *&0&0 \\
{}    && {}  & CD  & 16 & All & *&6&4   & *&1&0 & *&0&0 \\
\hline
\end{tabular}
}}
\end{table}

\smallskip
\noindent{\bf Description of tables.}
The caption of Table~\ref{tabConnect} describes the column headings used in Tables~\ref{tabConnect}-\ref{tabTTT}, 
which describe the experimental results. 
In our experiments, we explore vertices from $W_2$ and $W_3$ only as the set $W_4$ 
is almost always empty. 
The third column $j=2,3$ denotes whether we explore from $W_2$ or $W_3$.
For the classification of a board position, 
we run the game between the depth-$\ka$ vs the depth-$\kb$ strategy for~30 times.
If player~1 wins (i)~more than $\frac{2}{3}$ times (20 times), then it 
is identified as easy (E); (ii)~less than $\frac{1}{3}$ times (10 times), 
then it is identified as hard (H); (iii)~else as medium (M).

\smallskip
\noindent{\bf Experimental results for CONNECT games.}
Table~\ref{tabConnect}  presents results for CONNECT-3 and CONNECT-4 games, 
against depth-3 strategies of the opponent.
An interesting finding
is that in CONNECT-4 games 
with board size $5$$\times$$5$, for all winning conditions (RCD, RD, CD, RC), 
there are easy, medium, and hard vertices, for $\ka$$=$$1$,$2$, and $3$, when $j$$=$$3$.
In other words, even in much smaller board size ($5$$\times$$5$ as compared to
the traditional $7$$\times$$7$) we discover interesting starting positions 
for CONNECT-4 games and its simple variants.

\begin{figure*}[ht]
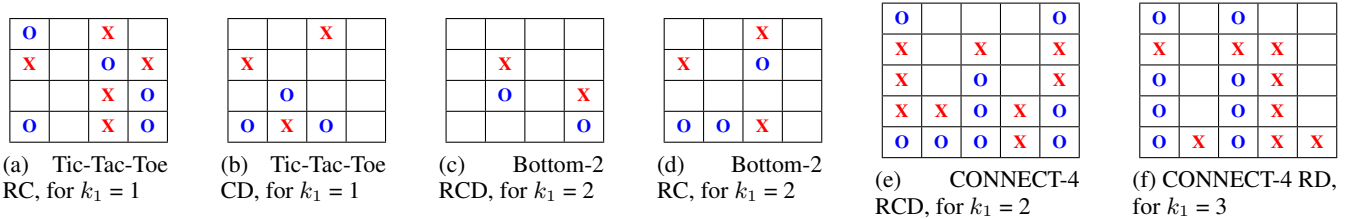
 
\centering
\subfloat[][\small Tic-Tac-Toe RC, for $\ka$ = 1]{
\bf
{\begin{tabular}{|C|C|C|C|} \hline
\OO &  & \XX &   \\ \hline
\XX &  & \OO & \XX \\ \hline
  &  & \XX & \OO \\ \hline
\OO &  & \XX & \OO \\ \hline
\end{tabular}\label{ex1}}}
\qquad
\subfloat[][\small Tic-Tac-Toe CD, for $\ka$ = 1]{
\bf
{\begin{tabular}{|C|C|C|C|} \hline
 & & \XX & \\ \hline
\XX & & & \\ \hline
 & \OO & & \\ \hline
\OO & \XX & \OO & \\ \hline
\end{tabular}}}
\qquad
\subfloat[][\small Bottom-2 RCD, for $\ka$ = 2]{
\bf
{\begin{tabular}{|C|C|C|C|} \hline
 & & & \\ \hline
 & \XX & & \\ \hline
 & \OO & & \XX \\ \hline
 & & & \OO \\ \hline
\end{tabular}}}
\qquad
\subfloat[][\small Bottom-2 RC, for $\ka$ = 2]{
\bf
{\begin{tabular}{|C|C|C|C|} \hline
 & & \XX & \\ \hline
\XX & & \OO & \\ \hline
 & & & \\ \hline
\OO & \OO & \XX & \\ \hline
\end{tabular}}}
\qquad
\subfloat[][\small CONNECT-4 RCD, for $\ka$ = 2]{
\bf
{\begin{tabular}{|C|C|C|C|C|} \hline
\OO & & & & \OO \\ \hline
\XX & & \XX & & \XX \\ \hline
\XX & & \OO & & \XX \\ \hline
\XX & \XX & \OO & \XX & \OO \\ \hline
\OO & \OO & \OO & \XX & \OO \\ \hline 
\end{tabular}}}
\qquad
\subfloat[][\small CONNECT-4 RD, for $\ka$ = 3]{
\bf
{\begin{tabular}{|C|C|C|C|C|} \hline
\OO & & \OO & & \\ \hline
\XX & & \XX & \XX & \\ \hline
\OO & & \OO & \XX & \\ \hline
\OO & & \OO & \XX & \\ \hline
\OO & \XX & \OO & \XX & \XX \\ \hline
\end{tabular}\label{ex6}}} 
\caption{Some ``Hard'' starting board positions generated by our tool for a variety of games and a different expertise level $\ka$ of player~1.
The opponent expertise level $\kb$ is set to $3$. Player~1 ({\XX}) can win in 2 steps for games (a)-(e) and in 3 steps for (f).
}
\end{figure*}

\smallskip
\noindent{\bf Experimental results for Bottom-2 games.}
Table~\ref{tabBottom2} shows the results for Bottom-2 (partial gravity-$2$) against  
depth-3 strategies of the opponent.
In contrast to CONNECT games, 
medium or hard vertices do not exist
for depth-3 strategies. 

\smallskip
\noindent{\bf Experimental results for Tic-Tac-Toe games.}
The results for Tic-Tac-Toe games are shown in Table~\ref{tabTTT}. 
For Tic-Tac-Toe games the strategy exploration is expensive (a tree of depth-3
for $4\times 4$ requires exploration of $10^6$ nodes). 
Hence using the iterative simulation techniques we first assign a score to all
vertices and use exploration for bottom hundred vertices (B100), i.e., hundred vertices
with the least score according to our iterative simulation metric.
In contrast to CONNECT games, 
interesting vertices exist only for depth-1 strategies.


\begin{table}
\centering
{\scriptsize
\caption{\small Tic-Tac-Toe against depth-$3$ strategy of opponent.
The sampling B100 denotes exploring vertices with the least scored hundred
vertices according to iterative simulation score.}
\label{tabTTT} %
\scalebox{0.96}{
\begin{tabular}{|*{6}c|*{3}c|*{3}c|*{3}c|} \hline
Game & State & $j$ & {Win} & {No. of} & Sampling & \multicolumn{9}{c|}{$\kb=3$}   \\ \cline{7-15}
{} & {Space} & {} & {Cond} & {States} & {} & \multicolumn{3}{c|}{$\ka$  = 1} & \multicolumn{3}{c|}{$\ka$  = 2} & \multicolumn{3}{c|}{$\ka$  = 3} \\ 
{} & {$|V|$} & &  & {$|W_j|$} & {} & E&M&H & E&M&H & E&M&H  \\ \hline

{3x3} & {5.4\e3} & $2$ & RCD & 36 & All & *&14&2 & *&0&0 & *&0&0 \\ 
{} 	  & {5.6\e3} & {}  & RC  & 0  & -   &  &&   &   &&  &   &&  \\ 
{}    & {5.6\e3} & {}  & RD,CD  & 1  & All & *&0&0  & *&0&0 & *&0&0 \\ 

{3x3} & & $3$ & Any & 0 & - & &&  & && & &&  \\ \hline

{4x4} & {6.0\e6} & $2$ & RCD & 128 & All & *&6&2  & *&0&0 & *&0&0 \\   
{} & {7.2\e6} {} & {} & RC & 3272 & B100 & *&47&22 & *&0&0 & *&0&0 \\ 
{} & {7.2\e6} {} & {} & RD,CD & 4627 & B100 & *&3&2 & *&0&0 & *&0&0 \\ 

{4x4} &    & $3$ & RCD, RC 	& 0 & -   &  &&  & && & && \\ 
{} 	  & {} & {}  & RD,CD 		& 4 & All & *&0&0 & *&0&0 & *&0&0 \\
\hline
\end{tabular}
}
}
\end{table}

\smallskip\noindent{\bf Running times.}
The generation of $W_j$ for $j$=$2$,$3$ took between $2$-$4$ hours
per game (this is a one-time computation for each game).
The time to classify a vertex as E, M, or H for depth-3 strategies of 
both players, playing 30 times from a board position on average varies between 
12 sec.~(for CONNECT-4 games) to 25 min.~for Tic-Tac-Toe games.
Details for depth-2 strategy of the opponent are given in the appendix.

\smallskip
\noindent{\bf Important findings.} 
Our first key finding is the \emph{existence of vertices of different hardness
levels} in various games.
We observe that in Tic-Tac-Toe games only board positions that are hard
for $\ka=1$ exist; in particular, and very interestingly, they also exist in board of size 
$4\times 4$.
Since the default start (the blank) vertex in $4 \times 4$ Tic-Tac-Toe games 
is heavily biased towards the player who starts first, they have been believed
to be uninteresting for ages, whereas our experiments discover interesting 
starting vertices for them.
With the slight variation of allowable moves (Bottom-2), we obtain board 
positions that are hard for $\ka=2$.
In Connect-4 we obtain vertices that are hard for $\ka=3$ even with small 
board size of $5\times 5$. 
The default starting vertex in Tic-Tac-Toe $3\times 3$ and Connect-4 $5\times 5$ does not belong to the winning set $W_j$; in Tic-Tac-Toe $4\times 4$ it belongs to the winning set $W_j$ and is Easy for all depth strategies. 

The second key finding of our results is that {\em the number of interesting 
vertices is a negligible fraction of the huge state space}. 
For example, in Bottom-2 RCD games with board size $4\times 4$ the size of
the state space is over $1.8$ million, but has only two positions 
that are hard for $\ka=2$; 
and in CONNECT-4 RCD games with board size $5\times 5$ the state space size is 
around sixty nine million, but has around two hundred hard vertices for 
$\ka=3$ and $\kb=3$, when $j=3$, among the five thousand vertices sampled 
from $W_j$.
Since the size of $W_j$ in this case is around $2.8 \times 10^5$, the total
number of hard vertices is around twelve thousand (among sixty nine million 
state space size).
Since the interesting positions are quite rare, a naive approach of 
randomly generating positions and measuring its hardness will be 
searching for a needle in a haystack and be ineffective 
to generate interesting positions. 
Thus there is need for a non-trivial search strategy (\sectref{methodology}), which our tool implements.

\smallskip
\noindent{\bf Example board positions.} 
In Figure~1(a)-Figure~1(f) we present examples of several board positions 
that are of different hardness level for strategies of certain 
depth. 
Also see \ifciteTech{appendix (attached as supplementary material) 
}
\else{appendix}
\fi for an illustration.
In all the figures, player-X is the current player against opponent of depth-$3$ strategy.
All these board positions were discovered through our experiments.

\section{Related Work} 
\label{sec-related}

\noindent{\bf Tic-Tac-Toe and Connect-4.}
Tic-Tac-Toe has been generalized to different board sizes, 
match length~\cite{weijima}, and even polyomino matches~\cite{Harary} 
to find variants that are interesting
from the default start state. Existing research has focussed on
establishing which of these games have a winning
strategy~\cite{gardner1979mathematical,gardner1983,eric}. In contrast,
we show that even simpler variants can be interesting if we start from
certain specific states. 
Connect-4 research has also focussed on establishing 
a winning strategy from the default starting state~\cite{allis1988}.
 In contrast, we study how easy or difficult is
to win from winning states given 
expertise levels. 

BDDs have been used to represent board games before \cite{kissmann2011gamer} to perform MCTS run with Upper Confidence Bounds applied to Trees (UCT). However, the goal was instantiation of the BDDs and finding the number of states explored by a single agent. In our setting we have two players, and use BDDs to compute the winning set. 

\smallskip
\noindent{\bf Level generation.}
\ignore{Our proposed technique, which generates starting states given certain parameters (namely, expertise levels of players, and hardness level), 
can be used to generate different game levels by simply varying the choice of the parameters along some partial/total order. Hence, we 
discuss some related work from the area of level generation.}
Goldspinner~\cite{aiide12} is a level generation system for KGoldrunner,
a 
puzzle game with dynamic elements. It
uses a genetic algorithm to generate candidate levels and simulation to 
evaluate dynamic aspects of the game. We also use simulation to evaluate
the dynamic aspect, but use symbolic methods to generate candidate states; 
also, our system is parameterized by game rules. 

Most other work has been restricted to games without opponent and
dynamic content such as Sudoku~\cite{sudoku07,sudoku09}. 
Smith et al. used answer-set programming to generate levels
for 
Refraction 
that adhered to pre-specified constraints written
in first-order logic~\cite{Smith12}. 
Similar approaches have also been
used to generate levels for platform games~\cite{Smith09}. 
In these approaches, designers must explicitly specify
constraints on the generated content, 
e.g., the tree needs to be near the rock and the river needs to
be near the tree. In contrast, our system takes as input rules of the
game and does not require any further help from the designer. 
\cite{chi13} also uses a similar model and 
applies symbolic methods (namely, test input generation
techniques) to generate various levels for DragonBox, 
which became the most purchased game in
Norway on 
the
 Apple App Store~\cite{Liu12}. In contrast, we use
symbolic methods for generating 
start states, and use simulation
for estimating their hardness level. 

\smallskip
\noindent{\bf Problem generation.}
Automatic generation of fresh problems can be a key capability in intelligent tutoring systems~\cite{cacm14}. The technique for generation of algebraic proof problems~\cite{aaai12} uses probabilistic testing to guarantee the validity of a 
generated problem candidate (from abstraction of the
original problem) on random inputs, but there is no guarantee of the
hardness level. Our simulation can be linked to this probabilistic
testing approach, but it is used to guarantee hardness level;
whereas validity is guaranteed by symbolic methods. 
The technique for generation of natural deduction problems~\cite{ijcai13:pgen} and~\cite{aaai14} involves a 
backward existential search over the state space of all possible proofs for all possible facts to dish out problems with a specific hardness level. In contrast, we employ a two-phased strategy of 
backward and forward search; backward search is necessary to identify winning states, while forward search ensures hardness levels. Furthermore, our state transitions alternate between different players, thereby necessitating alternate universal vs.~existential search over transitions.

Interesting starting states that require few steps to play and win are
often published in newspapers for sophisticated 
games like Chess and Bridge. 
These are usually obtained from database of past games. In contrast,
we show how to automatically generate such states, albeit for 
simpler games.

\ignore{Level generation is an important problem that has been studied for
games. In the context of 1-player games, such as Sudoko, MineSweeper,Dragon Box, or Refraction, it involves coming up with an appropriate 
initial state. In the context of 2-player games, such as Connect-4, level generation is
controlled only through selecting the expertise of the opponent, which
is played by the computer. In
this work, we show how to generate levels for 2-player games by
generating appropriate initial states. This has the advantage of
taking into account various important parameters such as number of
steps required to win and the expertise of the current player. 

In fact, the higher popularity of multi-player
computer games above solo computer games show how much we still enjoy
playing with other people than the computer.

Interesting starting states that require few steps to play and win are
often published in newspapers for sophisticated 
games like Chess and Bridge. 
These are usually obtained from database of past games. In contrast,
we show how to automatically generate such states, albeit for 
simpler games.}

{\small
\smallskip\noindent{\bf Acknowledgments.}
The research was partly supported by Austrian Science Fund (FWF) Grant No P23499- N23, 
FWF NFN Grant No S11407-N23 (RiSE), ERC Start grant (279307: Graph Games), and Microsoft faculty fellows award.

\bibliographystyle{aaai}
\balance
\bibliography{paper}

}

\appendix
\clearpage
\section{Appendix}

\smallskip\noindent{\bf Strategies.}
A strategy for a player is a rule that specifies how to extend plays.
Formally, a \emph{strategy} $\straa$ for player~1 is a function 
$\straa$: $V^* \cdot V_1 \to \distr(V)$, where $\distr(V)$ denote the set of 
probability distributions over $V$, that given a finite sequence of vertices 
(representing the history of the play so far) which ends in a player~1 
vertex, chooses a probability distribution for the next vertex.
The strategy must choose only available successors, i.e., for all $w \in V^*$ 
and $v \in V_1$ we have that if $\straa(w \cdot v)(u)>0$, then $(v,u)\in E$.
The strategies $\strab$ for player~2 are defined analogously.


\begin{table}[!htb]
\centering
{\scriptsize
\caption{\small CONNECT-3 \& -4 against depth-$2$ strategy of opponent;
(C-3 (resp. C-4) stands for CONNECT-3 (resp. CONNECT-4)).
The third column ($j$) denotes whether we explore from $W_2$ or $W_3$.
The sixth column denotes sampling to select starting vertices 
if $|W_j|$ is large: ``All'' denotes that we explore all states in $W_j$, 
and Rand denotes first sampling 5000 states randomly from $W_j$ and 
exploring them.
The E, M, and H columns give the number of easy, medium, or hard states among 
the sampled states.
For each $\ka=1,2,$ and $3$ the sum of E, M, and H columns is equal
to the number of sampled states, and * denotes the number of remaining states.
Observe that $|W_j|$ is small fraction of $|V|$ (this illustrates the significance of our use of symbolic methods
as opposed to the prohibitive explicit enumerative search). Also, observe that states labeled medium and hard are a small 
fraction of the sampled states (this illustrates the significance of our efficient iterative sampling strategy).
} 
\label{tabConnect-2} 
\scalebox{0.8}{
\begin{tabular}{|*{6}c|*{3}c|*{3}c|*{3}c|} \hline
Game & State & $j$ & {Win} & {No. of} & Sampling & \multicolumn{9}{c|}{$\kb=2$}   \\ \cline{7-15}
{} & {Space} & {} & {Cond} & {States} & {} & \multicolumn{3}{c|}{$\ka$  = 1} & \multicolumn{3}{c|}{$\ka$  = 2} & \multicolumn{3}{c|}{$\ka$  = 3} \\ 
{} & {$|V|$} & &  & {$|W_j|$} & {} & E&M&H & E&M&H & E&M&H  \\ \hline

{C-3}	& {4.1\e4} & 2 & RCD & 110 & All & *&24&5  & *&3 &0  & *&0&0 \\ 
{4x4}	& {6.5\e4} &   & RC  & 200 & All & *&39&9  & *&23&5  & *&0&0 \\ 
{}		& {7.6\e4} &   & RD  & 418 & All & *&38&14 & *&24&4  & *&0&0 \\ 
{}		& {6.5\e4} &   & CD  & 277 & All & *&44&21 & *&27&17 & *&0&0 \\ \hline 

{C-3} & & 3 & RCD, RC, CD 	& 0  & -   & 	 &&& 	 &&&    &&  	 \\ 
{4x4} & &   & RD 			& 18 & All & *&0&0 & *&0&0 & *&0&0  \\ \hline

{C-4} 	& 6.9\e7  & 2 & RCD & 1.2\e6 & Rand & *&183&202 & *&148&115 & *&0&0 \\ 
{5x5} 	& {8.7\e7}&   & RC  & 1.6\e6 & Rand & *&70&237  & *&75&181  & *&0&0 \\ 
{}		& {1.0\e8}&   & RD  & 1.1\e6 & Rand & *&116&268 & *&144&77  & *&0&0 \\ 
{} 		& {9.5\e7}&   & CD  & 5.3\e5 & Rand & *&357&133 & *&200&95  & *&0&0 \\ \hline 

{C-4}&& 3 & RCD & 2.8\e5 & Rand & *&445&832  & *&384&497 & *&227&166 \\ 
{5x5}&&   & RC  & 7.7\e5 & Rand & *&328&969  & *&328&506 & *&93&196  \\
{}	 &&   & RD  & 8.0\e5 & Rand & *&398&1206 & *&477&501 & *&177&79  \\
{} 	 &&   & CD  & 1.5\e5 & Rand & *&146&73   & *&168&44  & *&87&19   \\ \hline
\end{tabular} 
}}
\end{table}

\begin{table}[!htb]
\centering
{\scriptsize
\vspace{-2em}
\caption{\small Bottom-2 against depth-$2$ strategy of opponent.} 
\label{tabBottom2-2} 
\scalebox{0.96}{
\begin{tabular}{|*{6}c|*{3}c|*{3}c|*{3}c|} \hline
Game & State & $j$ & {Win} & {No. of} & Sampling & \multicolumn{9}{c|}{$\kb=2$}   \\ \cline{7-15}
{} & {Space} & {} & {Cond} & {States} & {} & \multicolumn{3}{c|}{$\ka$  = 1} & \multicolumn{3}{c|}{$\ka$  = 2} & \multicolumn{3}{c|}{$\ka$  = 3} \\ 
{} & {$|V|$} & &  & {$|W_j|$} & {} & E&M&H & E&M&H & E&M&H  \\ \hline

{3x3} & {4.1\e3} & $2$ & RCD & 20 & All & *&5&0 & *&1&0 & *&0&0 \\ 
{} 	  & {4.3\e3} & {}  & RC  & 0  & -   &   &&  &    && &    && \\ 
{} 	  & {4.3\e3} & {}  & RD  & 9  & All & *&2&1 & *&3&0 & *&0&0 \\ 
{} 	  & {4.3\e3} & {}  & CD  & 1  & All & *&0&0 & *&0&0 & *&0&0 \\ \hline 

{3x3} &  & $3$ & Any & 0 & -  & &&  & && & && \\ \hline

{4x4} & {1.8\e6} & $2$ & RCD & 193  & All & *&14&25   & *&0&2    & *&0&0 \\ 
{} 	  & {2.4\e6} & {}  & RC  & 2709 & All & *&572&288 & *&89&245 & *&0&0 \\ 
{} 	  & {2.3\e6} & {}  & RD  & 2132 & All & *&104&48  & *&13&6   & *&0&0 \\ 
{} 	  & {2.4\e6} & {}  & CD  & 1469 & All & *&127&49  & *&18&5   & *&0&0 \\ \hline 

{4x4} && $3$ & RCD & 0  & -   &    &&   &   &&  &   &&  \\
{} 	  && {}  & RC  & 90 & All & *&38&27 & *&0&0 & *&0&0 \\
{}    && {}  & RD  & 24 & All & *&0&2   & *&0&0 & *&0&0 \\
{}    && {}  & CD  & 16 & All & *&6&3   & *&1&0 & *&0&0 \\
\hline
\end{tabular}
}}
\vspace{-2em}
\end{table}

\smallskip\noindent{\bf Summary and details of running time.}
We summarize our results in Table~\ref{tab4}.
We call a state \emph{category-$i$} state if it is not easy for 
depth-($i$$-$$1$) strategy, but it is easy for depth-$i$ strategy.
In Table~\ref{tab4} we summarize the different games and the existence
of category~$i$ states in such games.
The generation of $W_j$ for $j=2$ and $j=3$ took between two to four hours
per game (note that this is a one-time computation for each game).
The evaluation time to classify a state as E, M, or H is as follows:
for depth-3 strategies of both players, playing 30 times from a board 
position on average takes
(i)~12 seconds for CONNECT-4 games with board size $5\times 5$;
(ii)~47 seconds for Bottom-2 games with board size $4\times 4$;
and (iii)~25 minutes for Tic-Tac-Toe games with board size $4\times 4$.
Note that the size of the state space is around $10^8$ for CONNECT-4 games
with board size $5\times 5$, and testing the winning nature of a state takes at least few seconds. 
Hence an explicit enumeration of the state space cannot be used to obtain $W_j$ in reasonable time; in contrast, our 
symbolic methods  succeed to compute $W_j$ efficiently.


\begin{table}[!htb]
\centering
{\scriptsize
\caption{\small Tic-Tac-Toe against depth-$2$ strategy of opponent.
The sampling B100 denotes exploring states with the least scored hundred
states according to iterative simulation score.}
\label{tabTTT-2} 
\scalebox{0.96}{
\begin{tabular}{|*{6}c|*{3}c|*{3}c|*{3}c|} \hline
Game & State & $j$ & {Win} & {No. of} & Sampling & \multicolumn{9}{c|}{$\kb=2$}   \\ \cline{7-15}
{} & {Space} & {} & {Cond} & {States} & {} & \multicolumn{3}{c|}{$\ka$  = 1} & \multicolumn{3}{c|}{$\ka$  = 2} & \multicolumn{3}{c|}{$\ka$  = 3} \\ 
{} & {$|V|$} & &  & {$|W_j|$} & {} & E&M&H & E&M&H & E&M&H  \\ \hline

{3x3} & {5.4\e3} & $2$ & RCD & 36 & All & *&14&0 & *&0&0 & *&0&0  \\ 
{} 	  & {5.6\e3} & {}  & RC  & 0  & -   &   &&   &   &&  &   &&  \\ 
{}    & {5.6\e3} & {}  & RD  & 1  & All & *&0&0  & *&0&0 & *&0&0\\ 
{}    & {5.6\e3} & {}  & CD  & 1  & All & *&0&0  & *&0&0 & *&0&0\\ \hline 

{3x3} & & $3$ & Any & 0 & - & &&  & && & && \\ \hline

{4x4} & {6.0\e6} & $2$ & RCD & 128 & All & *&8&0   & *&0&0 & *&0&0  \\   
{} & {7.2\e6} {} & {} & RC & 3272 & B100 & *&48&21 & *&0&0 & *&0&0  \\ 
{} & {7.2\e6} {} & {} & RD & 4627 & B100 & *&3&2   & *&0&0 & *&0&0  \\ 
{} & {7.2\e6} {} & {} & CD & 4627 & B100 & *&3&2   & *&0&0 & *&0&0  \\ \hline 

{4x4} &    & $3$ & RCD, RC 	& 0 & -   &   &&  &   &&  & &&    \\ 
{} 	  & {} & {}  & RD 		& 4 & All & *&0&0 & *&0&0 & *&0&0 \\
{}    & {} & {}  & CD 		& 4 & All & *&0&0 & *&0&0 & *&0&0 \\ \hline
\end{tabular}
}
}
\vspace{1em}
\end{table}

\begin{table}[!htb]
\centering
{\tiny

\caption{Summary}\label{tab4} 
\scalebox{0.96}{

\begin{tabular}{|c|c|c|c|c|} \hline
Game & {Category-1} & {Category-2} & {Category-3} & {Category-4} \\ \hline
{Tic-Tac-Toe} & {All variants} & {3x3: only RCD} & {} & {} \\ 
{} & {} & {4x4: All $j$=2 variants} & {} & {} \\ \hline
{Bottom-2} & {All variants} & {3x3: only RD} & {4x4: All $j$=2 variants} & {} \\ 
{} & {} & {4x4: All variants} & {} & {} \\ \hline
{CONNECT-3} & {All variants} & {All $j$=2 variants} & {All $j$=2 variants except RCD} & {} \\ \hline 
{CONNECT-4} & {All variants} & {All variants} & {All variants} & {All $j$=3 variants} \\ 
\hline\end{tabular}}
}
\end{table}

\begin{figure*}[t]{ \label{ttt_1}
\centering
\includegraphics[scale=0.7]{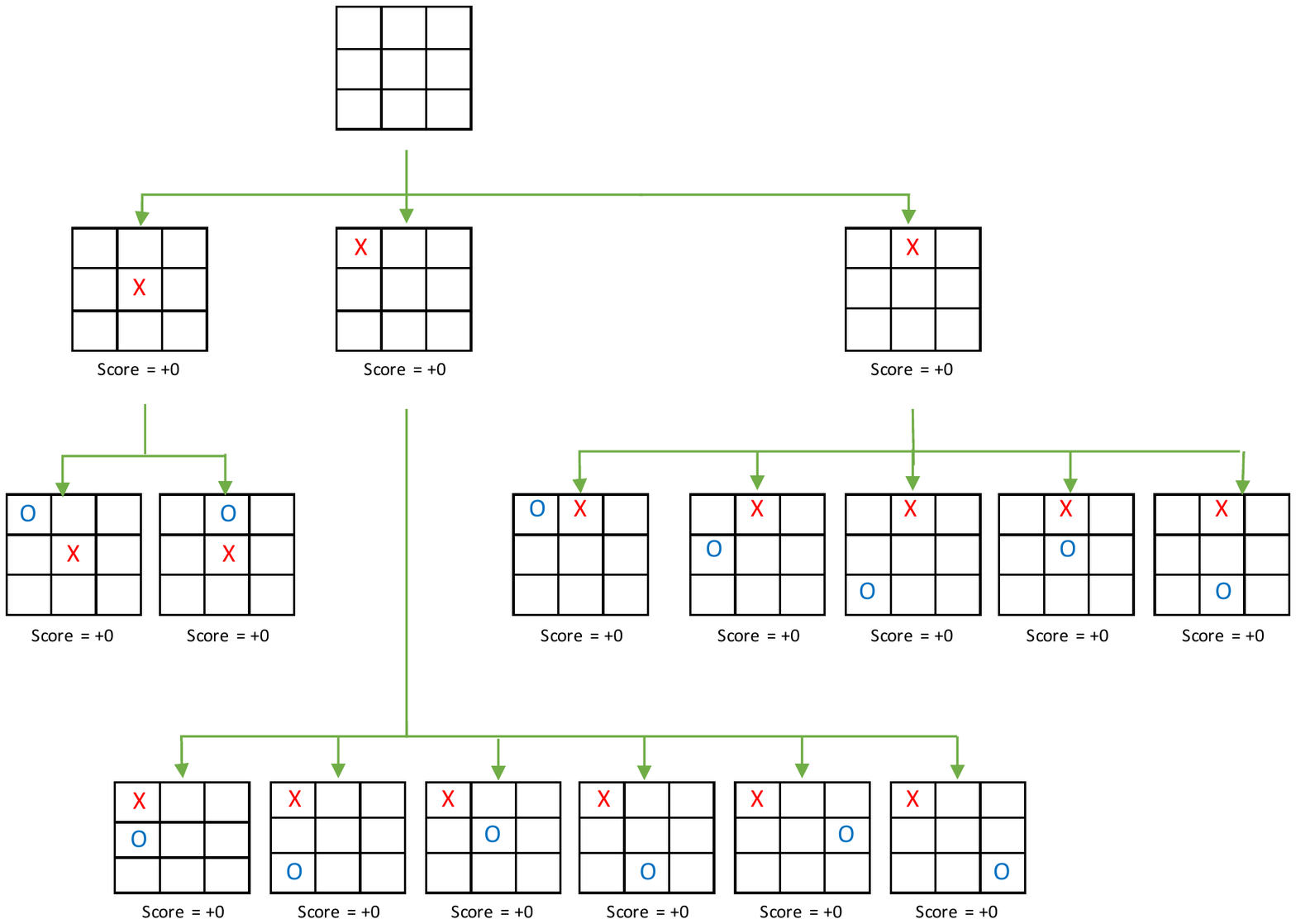} \hfill

}
\vspace{-23em}
\caption{Illustration of depth $\ka=1$ tree exploration for Tic-Tac-Toe. Since all possible moves for $\ka=1$ have equal score of $+0$, all board positions are chosen at random.}
\end{figure*}

\begin{figure*}{ \label{ttt_2}
\centering
\includegraphics[scale=0.7]{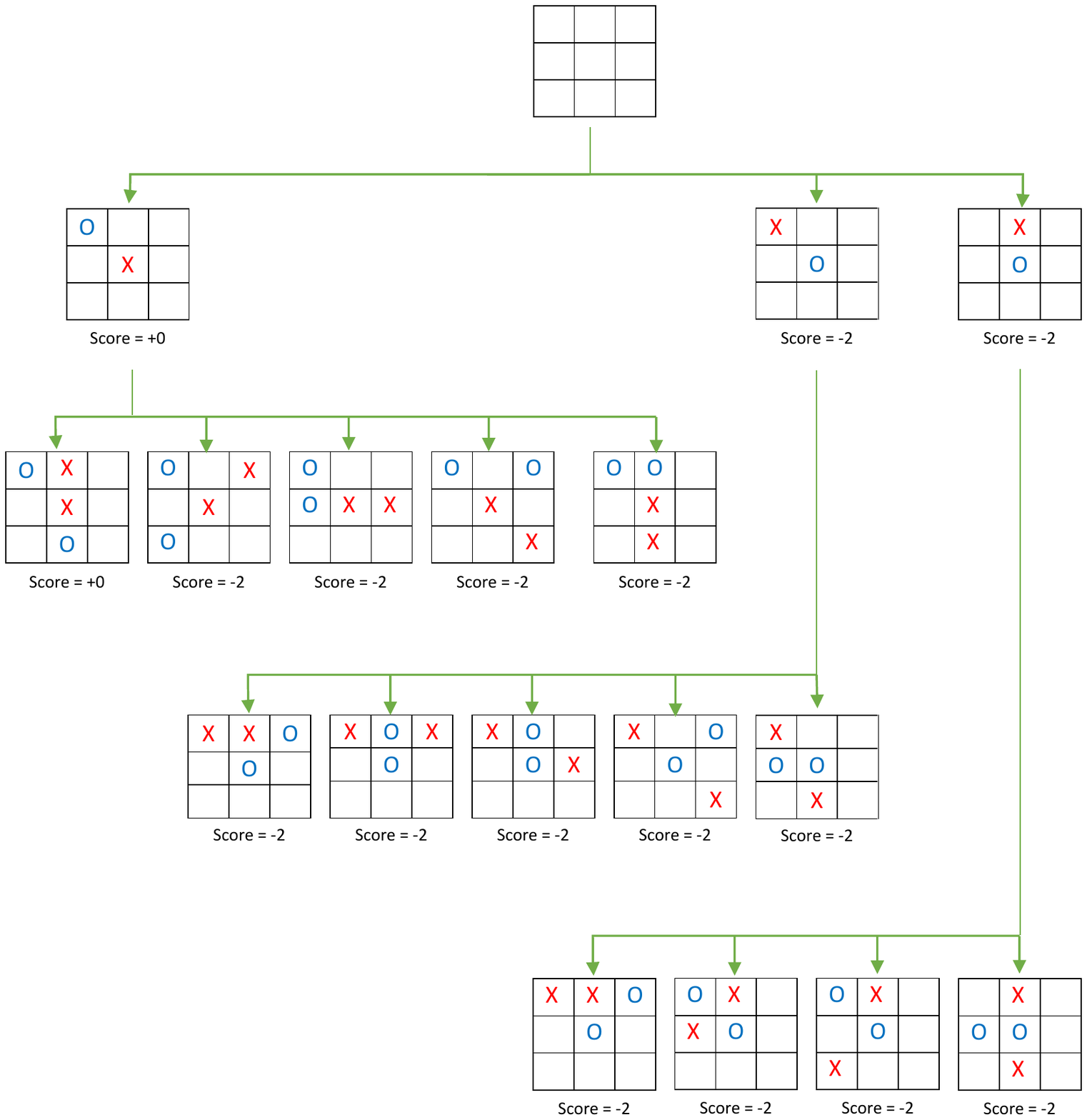}}
\vspace{-20em}
\caption{Illustration of depth $\ka=2$ tree exploration for Tic-Tac-Toe. In the figure, every choice of player~1 is followed by the optimal choice of opponent and they are collapsed to a single state. Since the center position has a higher score of $+0$, the $\ka=2$ strategy always chooses this move and considers all other positions to be equal with score of $-2$.}
\end{figure*}

\begin{figure*}
\centering
\epsfig{file=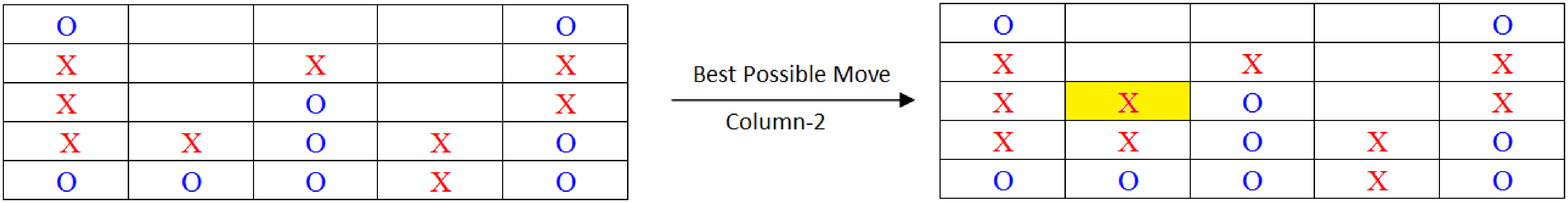, width=\textwidth}

\subfloat[$\ka=1$, column-2 fetches Player-X a reward of +5]{\epsfig{file=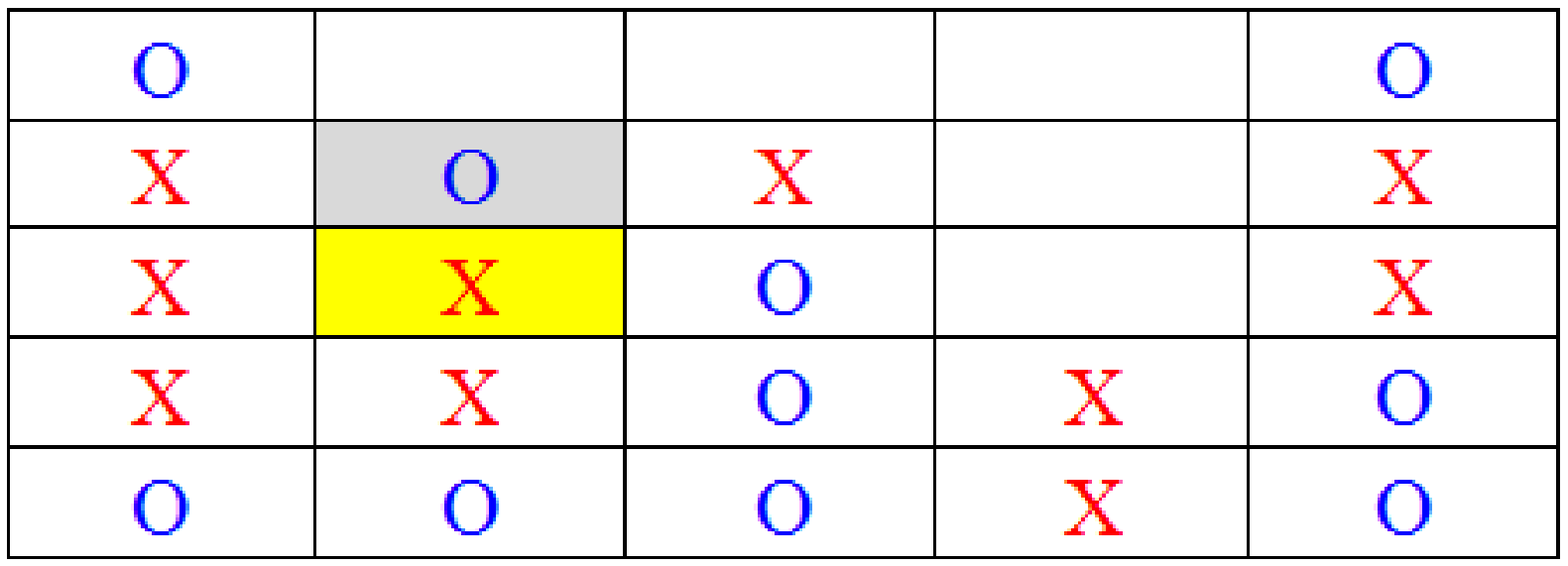, width=0.29\textwidth}}\qquad
\subfloat[$\ka=1$, column-3 fetches Player-X a reward of +6]{\epsfig{file=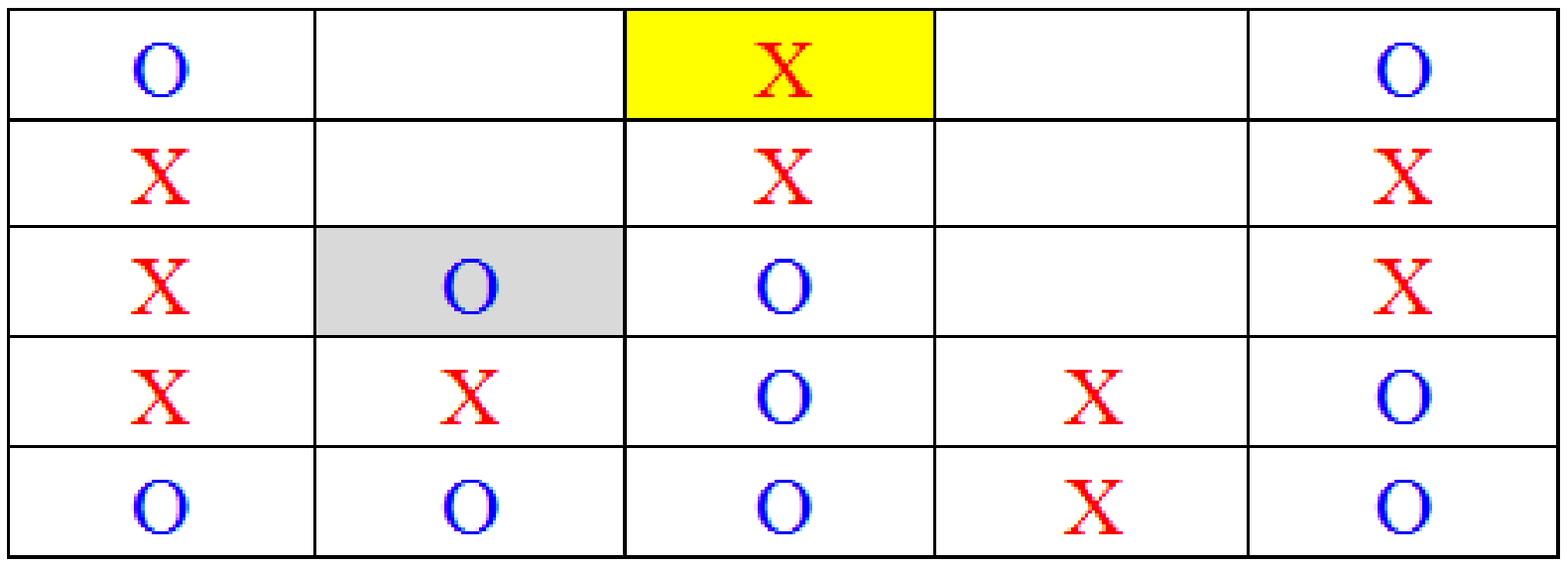, width=0.29\textwidth}}\qquad
\subfloat[$\ka=1$, column-4 fetches Player-X a reward of +2]{\epsfig{file=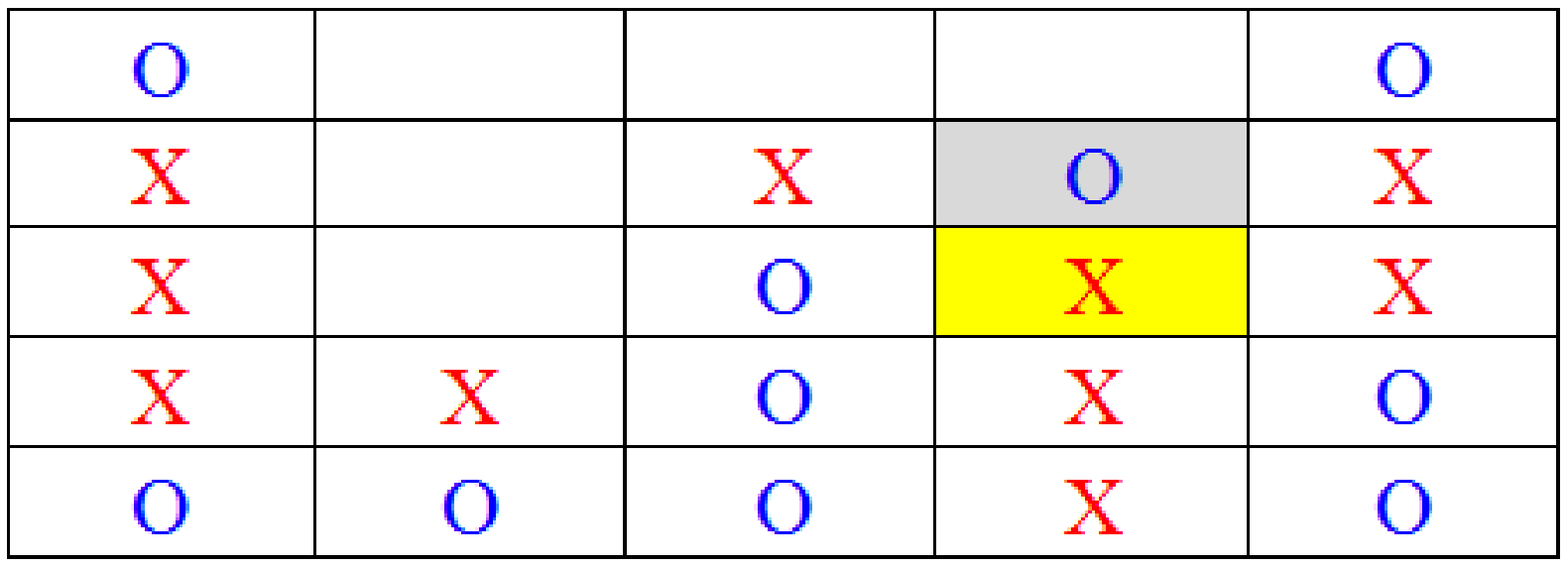, width=0.29\textwidth}}

\subfloat[$\ka=2$, column-2 fetches Player-X a reward of +3]{\epsfig{file=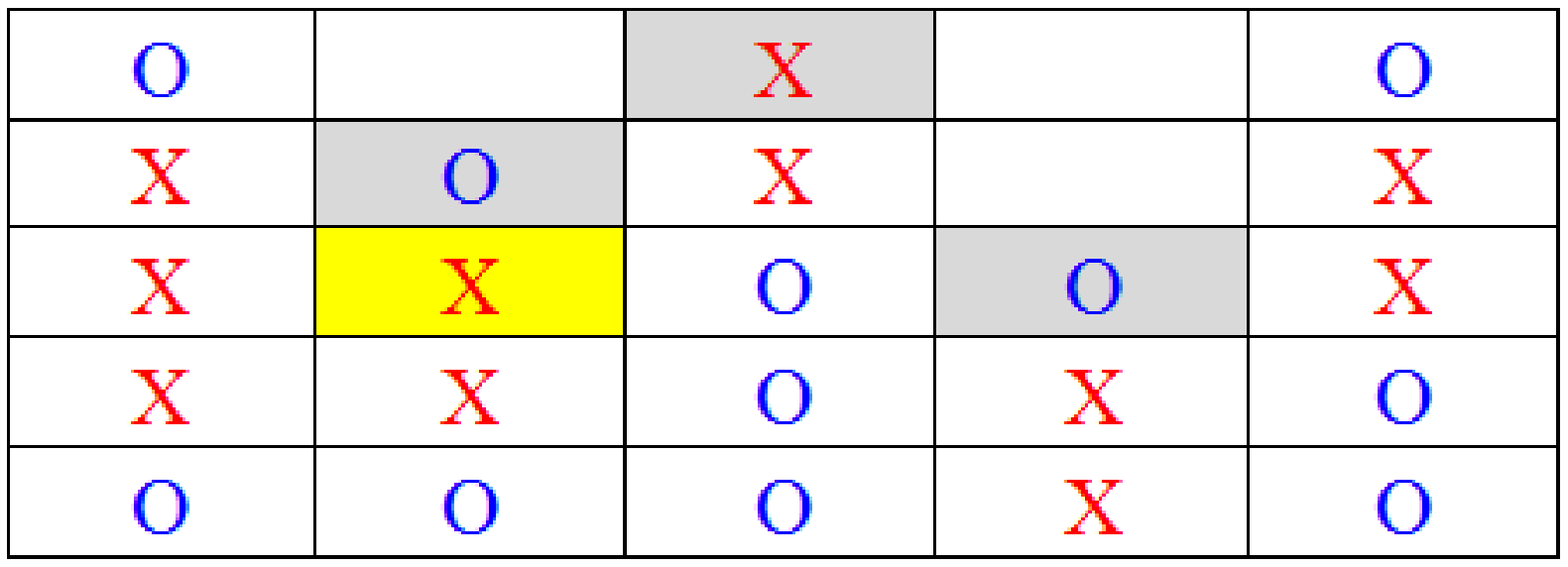, width=0.29\textwidth}}\qquad
\subfloat[$\ka=2$, column-3 fetches Player-X a reward of +6]{\epsfig{file=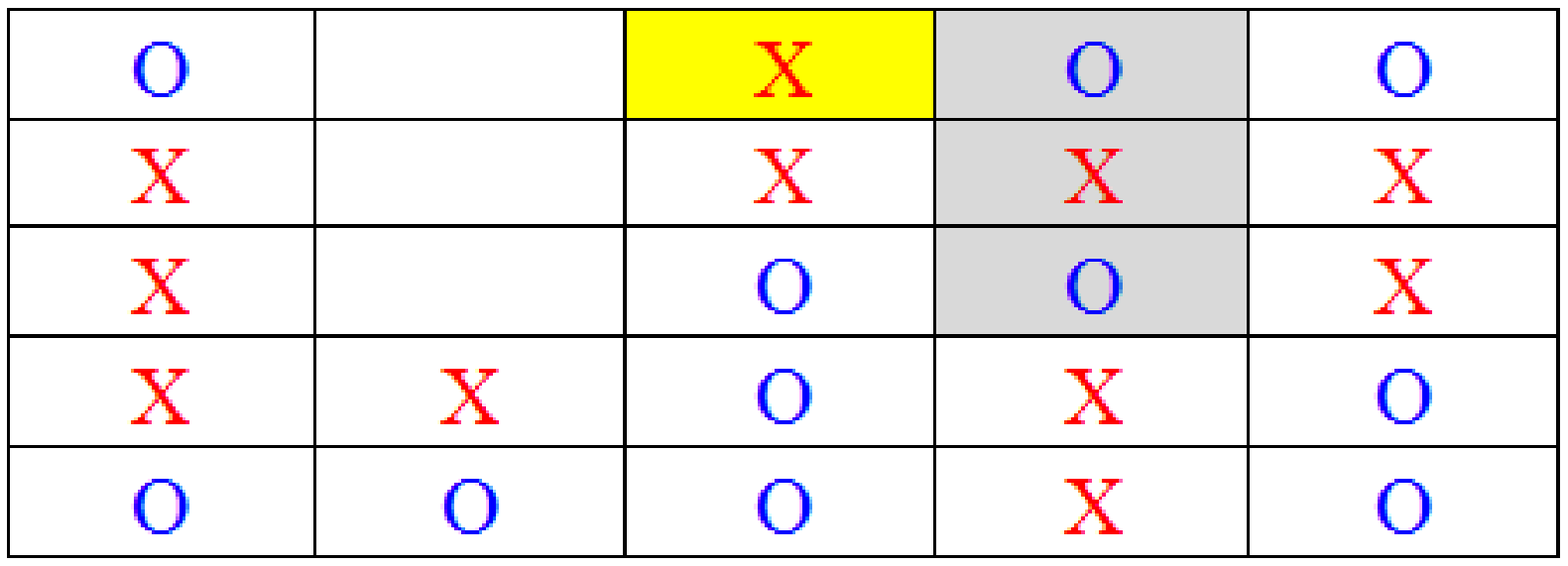, width=0.29\textwidth}}\qquad
\subfloat[$\ka=2$, column-4 fetches Player-X a reward of +0]{\epsfig{file=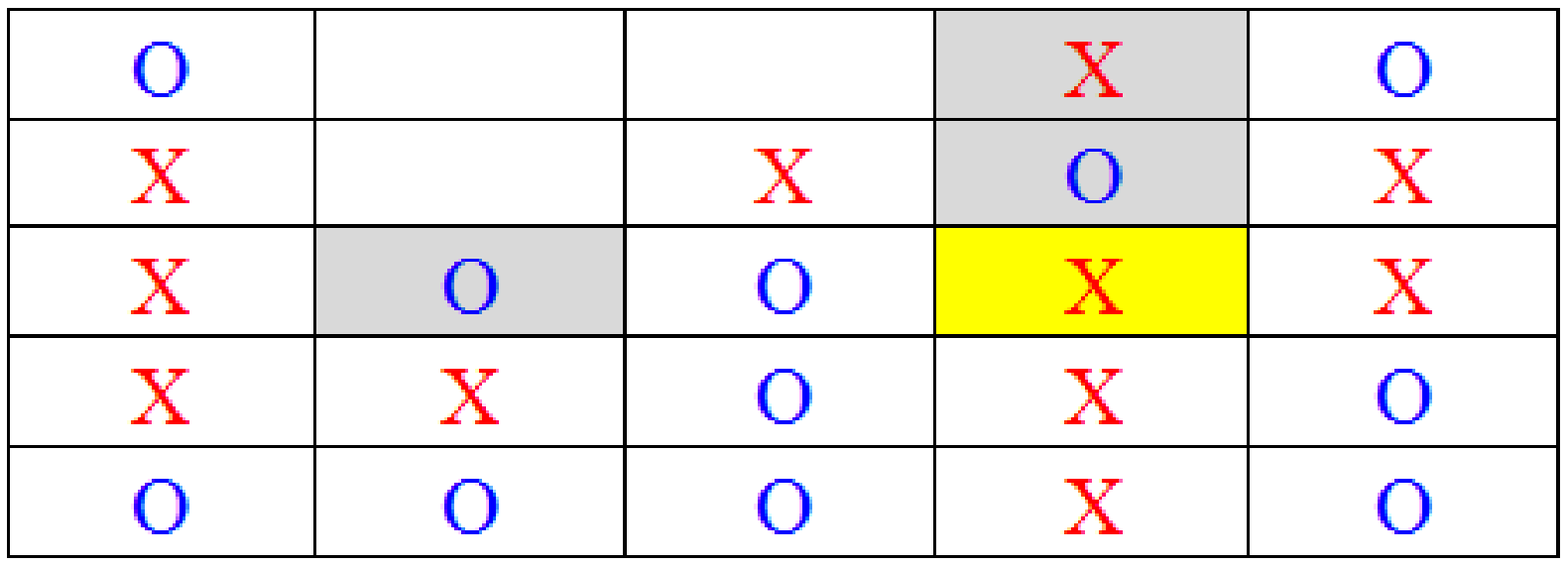, width=0.29\textwidth}}

\subfloat[$\ka=3$, column-2 fetches Player-X a reward of +$\infty$]{\epsfig{file=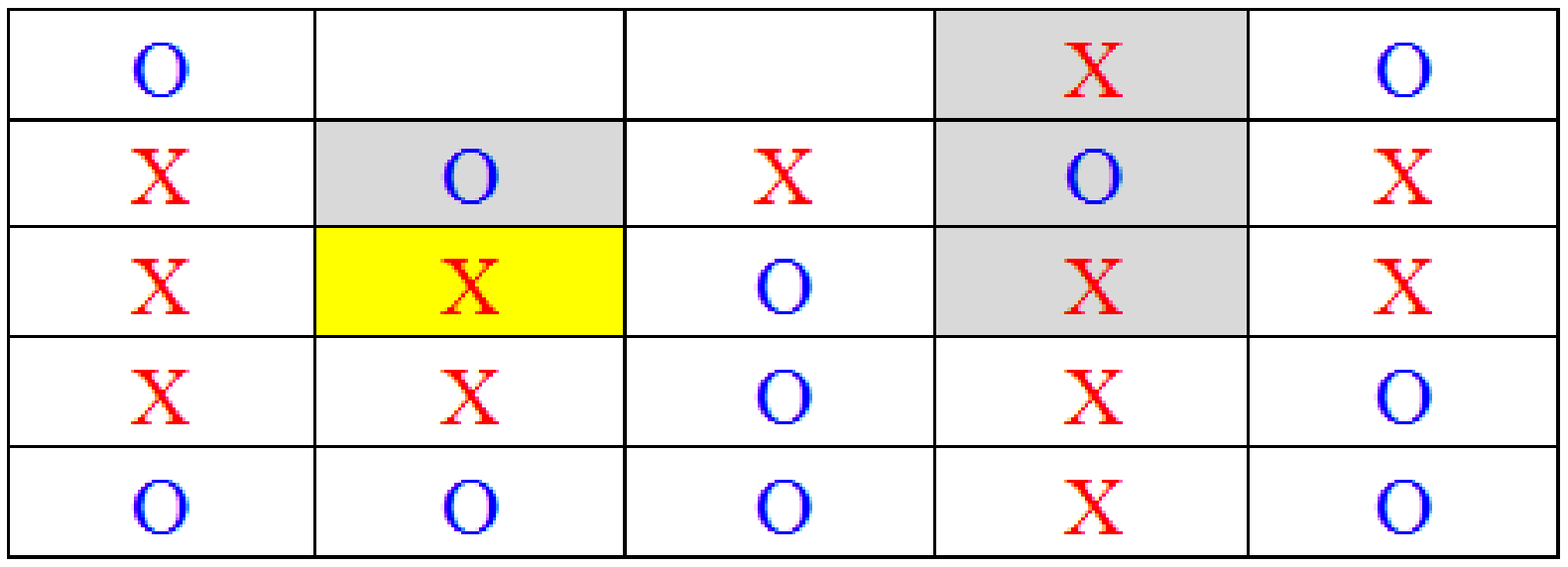, width=0.3\textwidth}}\qquad
\subfloat[$\ka=3$, column-3 fetches Player-X a reward of +0]{\epsfig{file=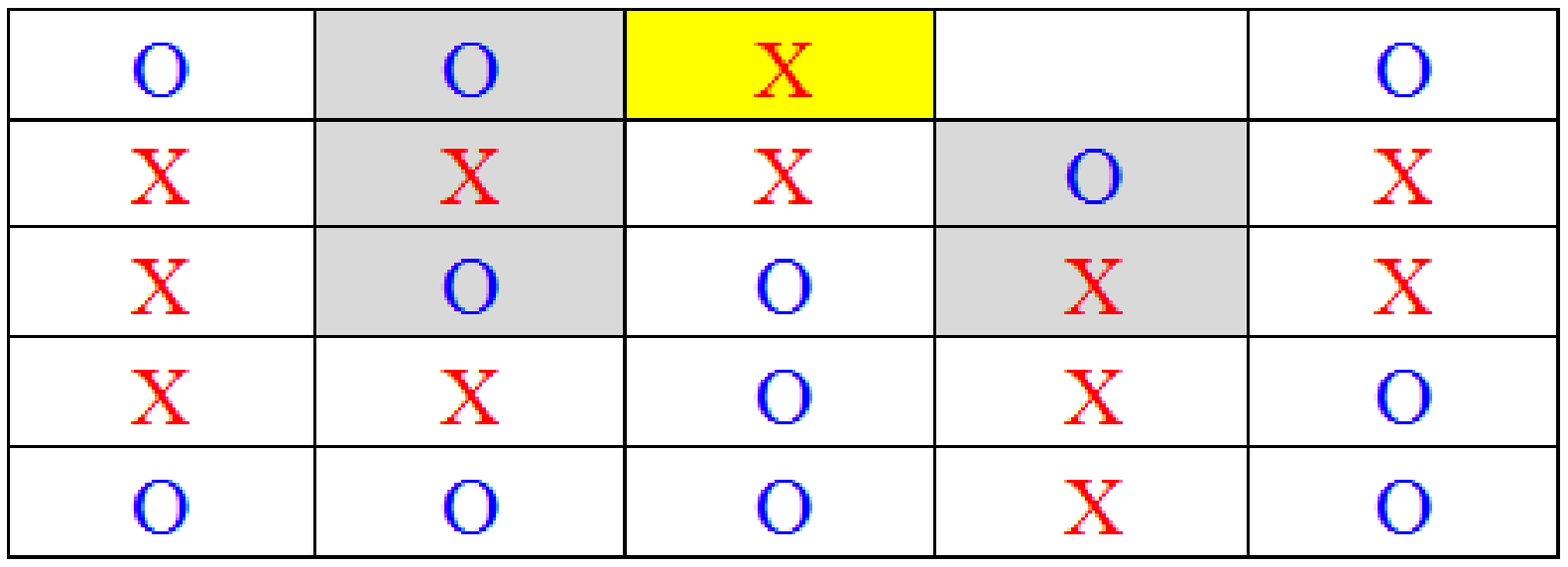, width=0.29\textwidth}}\qquad
\subfloat[$\ka=3$, column-4 fetches Player-X a reward of +0]{\epsfig{file=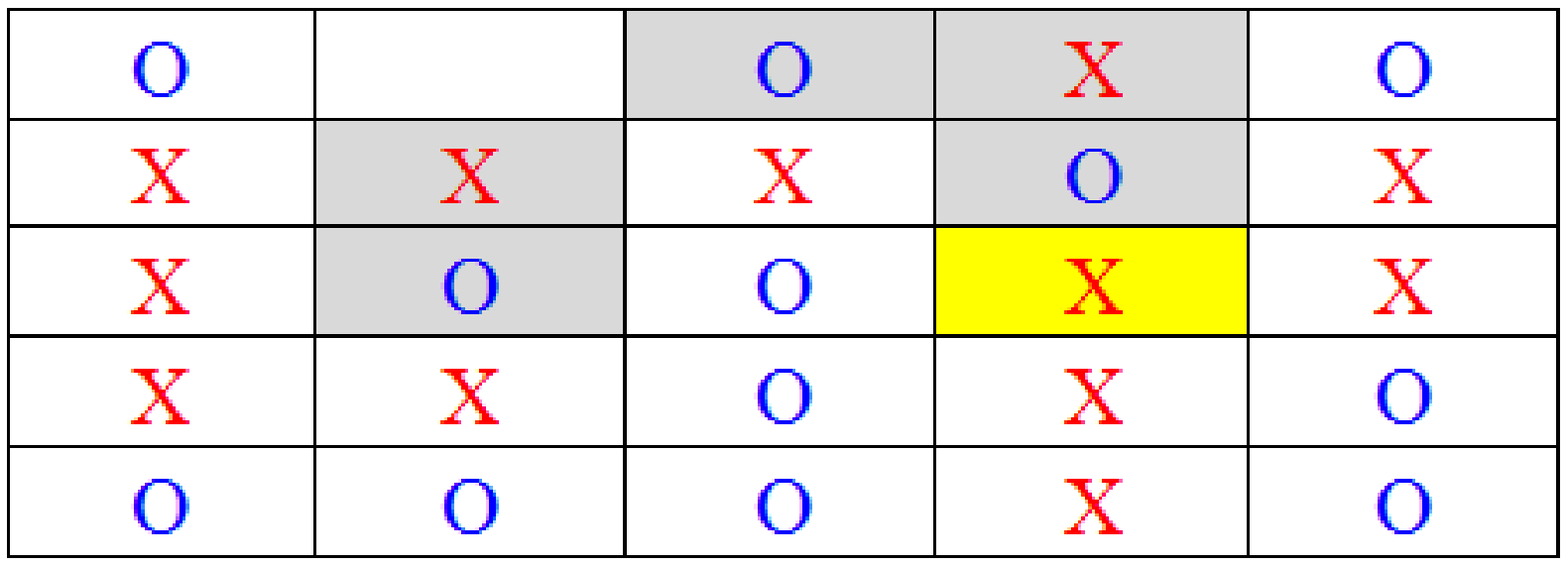, width=0.29\textwidth}}

\caption{Illustration of depth-$\ka$ strategy exploration on a CONNECT-4 RCD category-3 state.
The figure shows how different depth-$\ka$ strategies 
choose the best available position to mark on a Connect-4 RCD category-3 state.
The example board position of the figure is the same as 
for Figure~1~(e). 
The three depth-$\ka$ strategies ($\ka=1,2,3$) play as player-X and assign a score 
to each of the three available positions (column-$2,3,4$) by looking $\ka$-turns 
ahead. 
In each sub-figure, the position with yellow-background is the one chosen for 
exploration and the positions with grey-background are the predicted moves of 
how the game might turn out after $\ka$-turns.  
As observed, only $\ka=3$ strategy is able to foresee that marking column-$2$ 
would lead player-X to a winning state and also conclude that the other column 
choices will lead to a draw. Where as, $\ka=1,2$ incorrectly choose column-$3$ 
as the best position to mark hence making this starting position a category-$3$ 
state.
}\label{fig:illustrate}
\end{figure*}


\end{document}